\documentclass[letterpaper, 10 pt, conference]{ieeeconf}
\IEEEoverridecommandlockouts
\usepackage{calc}
\usepackage{url}
\usepackage{hyperref}
\usepackage{graphicx}
\usepackage[cmex10]{amsmath}
\usepackage{amssymb}
\usepackage{rotating}

\usepackage{nicefrac}
\usepackage{cite}
\usepackage[caption=false,font=footnotesize]{subfig}
\usepackage[usenames, dvipsnames]{color}
\usepackage{colortbl}
\usepackage{overpic}
\graphicspath{{./pictures/pdf/},{./pictures/ps/},{./pictures/png/},{./pictures/jpg/}}
\usepackage{breqn} %for breaking equations automatically
\usepackage[ruled]{algorithm}
\usepackage{algpseudocode}
\usepackage{multirow}

\newcommand{\figwid}{0.22\columnwidth}

\begin{document}

%%%%%%%%%%%%%% For debugging purposes, I like to display the TOC
%    \tableofcontents
%    \setcounter{tocdepth}{3}
%\newpage
%\mbox{}
%\newpage
%\mbox{}
%\newpage

%%%%%% END TOC %%%%%%%%%%%%%%%%%%%%%%%%%%%%%%%%%%%%%%%

\title{\LARGE \bf 
Crowdsourcing Swarm Manipulation Experiments:\\
A Massive Online User Study with Large Swarms of Simple Robots
%Crowdsourcing Swarm Manipulation:
%A massive online user study manipulation with large swarms of simple robots
%Massive Online Open eXperiments: \\ %Swarm Manipulation:
%Crowdsourcing Manipulation Experiments with Large Swarms of Simple Robots
%Online Massive Manipulation:\\ A massive user study on controlling massive swarms of simple robots
%Online Massive Manipulation:\\ A massive user study on controlling massive swarms of simple robots
}
\author{Aaron Becker, Chris Ertel, and James McLurkin%, 
\thanks{{A. Becker  is with the Department of Cardiovascular Surgery,  Boston Children's Hospital and Harvard Medical School, Boston, MA, 02115 USA {\tt\small aaron.becker@childrens.harvard.edu}, C. Ertel, and J. McLurkin are with the Computer Science Department, Rice University, Houston, TX 77005 USA  {\tt\small  \{cre1, jmclurkin\}@rice.edu}
}
} %\end thanks
} % end author block
\maketitle

\begin{abstract}
Micro- and nanorobotics have the potential to revolutionize many applications including targeted material delivery, assembly, and surgery.  The same properties that promise breakthrough solutions---small size and large populations---present unique challenges to generating controlled motion. We want to use large swarms of robots to perform manipulation tasks; unfortunately, human-swarm interaction studies as conducted today are limited in sample size, are difficult to reproduce, and are prone to hardware failures. We present an alternative.

This paper examines the perils, pitfalls, and possibilities we discovered by launching \href{http://www.swarmcontrol.net}{SwarmControl.net}, an online game where players steer swarms of up to 500 robots to complete manipulation challenges. We record statistics from thousands of players, and use the game to explore aspects of large-population robot control. We present the game framework as a new, open-source tool for large-scale user experiments. Our results have potential applications in human control of micro- and nanorobots, supply insight for automatic controllers, and provide a template for large online robotic research experiments.

%{The introduction tells us the contribution here.  What did we learn?}
  
% Control of multi-robot systems with large populations is in its infancy, and would benefit from detailed user experiments.  Such experiments could help select the best control architectures and level of feedback and lead ultimately to efficient automatic controllers.
% Unfortunately, large-scale user experiments have been prohibitively expensive.
   
%We present a survey of several control techniques for directing swarms,
%and compare their efficacy under several different task scenarios. The
%control rules used are either control strategies in the global frame,
%strategies which depend on user input and function in a local frame, or
%hybrid strategies that take into account both global frames and swarm state
%These rules are tested for their utility in swarm positioning, in manipulating
%objects through a maze, and in assembling objects.

\end{abstract}

%%%%%%%%%%% PAPER OUTLINE
% INTRODUCTION
%   micro/nano applications
%	open questions in swarm control
% 	difficulties in obtaining experimental data
% Related work
%%  HSI work
%%  Global control work
%%  online gamification
% Experimental Methods
%%  Platform  
%%  Human Subjects
%%% recruited through social media
%%% IRB form  Protocol Number: 14-012E Protocol Title: Massive Manipulation: A n online user study on controlling large swarms of simple robot sApproval Date: 7/26/2013Expiration Date: 7/26/2014
%%% Costs  for experiment:  ??
%%% Instrumenting:
%%%% Google analytics, airbrake, etc.
% RESULTS
%% EXPERIMENTAL RESULTS For each task, list the hypothesis, then give plot
%% DEMOGRAPHIC data
% Discussion and Future Work
%% Lessons learned
%%  Things that went wrong:   categorize what you can change in experiments without changing results (robot speed bad, color changes fine)
%% Color change for colorblind
%% not controlling for browser/computer speed.  Show statistics of what browsers used
% Conclusion 
%%  control with nonuniform flow
%%  demographics questionaires: 

%%%%%%%%%%%%%%%
\section{Introduction}\label{sec:Intro}
%This project studies system models and user interfaces for five multi-robot manipulation tasks with large populations of micro- and nanorobots.  We test several system models with different limitations on controllability and observability of the motion controller, and evaluate several different user interfaces.  We conduct user experiments to understand the impact of these limitations and design choices. 

%Micro- and nanorobotics have the potential to revolutionize many applications including targeted material delivery, assembly, and surgery.  The same properties that promise breakthrough solutions---small size and large populations---present unique challenges to generating controlled motion.  
Large populations of micro- and nanorobots are being produced in laboratories around the world, with diverse potential applications in drug delivery and construction \cite{Peyer2013,Shirai2005,Chiang2011}. These activities require robots that behave intelligently.
Limited computation and communication rules out autonomous operation or direct control over individual units; instead we must rely on global control signals broadcast to the entire robot population.  It is not always practical to gather pose information on individual robots for feedback control; the robots might be difficult or impossible to sense individually due to their size and location. However, it is often possible to sense global properties of the group, such as mean position and density.  Finally, many promising applications will require direct human control, but user interfaces to thousands---or millions---of robots is a daunting human-swarm interaction (HSI) challenge. 

The goal of this work is to provide a tool for investigating HSI methods through statistically significant numbers of experiments.  There is currently no comprehensive understanding of user interfaces for controlling multi-robot systems with massive populations.  We are particularly motivated by the sharp constraints in micro- and nanorobotic systems.  For example, full-state feedback with $10^6$ robots leads to operator overload.  
Similarly, the user interaction required to individually control each robot scales linearly with robot population.   
Instead, user interaction is often constrained to modifying a global input. This input may be nonstandard, such as the attraction/repulsion field from a scanning tunneling microscope (STM) tip. 

Our previous work with over a hundred hardware robots and thousands of simulated robots~\cite{Becker2013b} demonstrated that direct human control of large swarms is possible. Unfortunately, the logistical challenges of repeated experiments with over one hundred robots prevented large-scale tests.

\begin{figure}
\renewcommand{\figwid}{0.32\columnwidth}
\subfloat[][Vary Number]{\label{fig:VaryNum}
\begin{overpic}[width =\figwid]{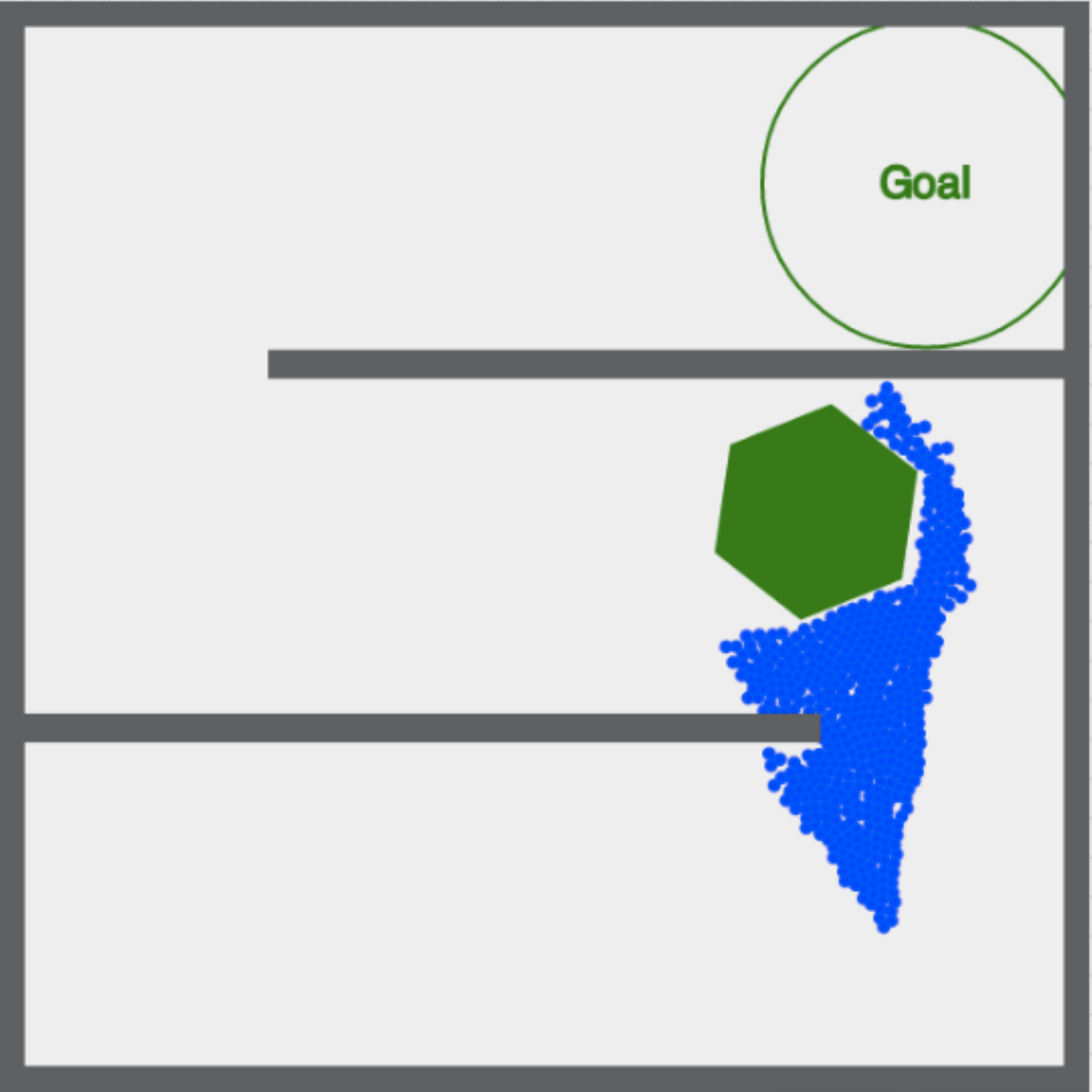}\end{overpic}}
\subfloat[][Vary Visual Feedback]{\label{fig:VaryVis}
\begin{overpic}[width =\figwid]{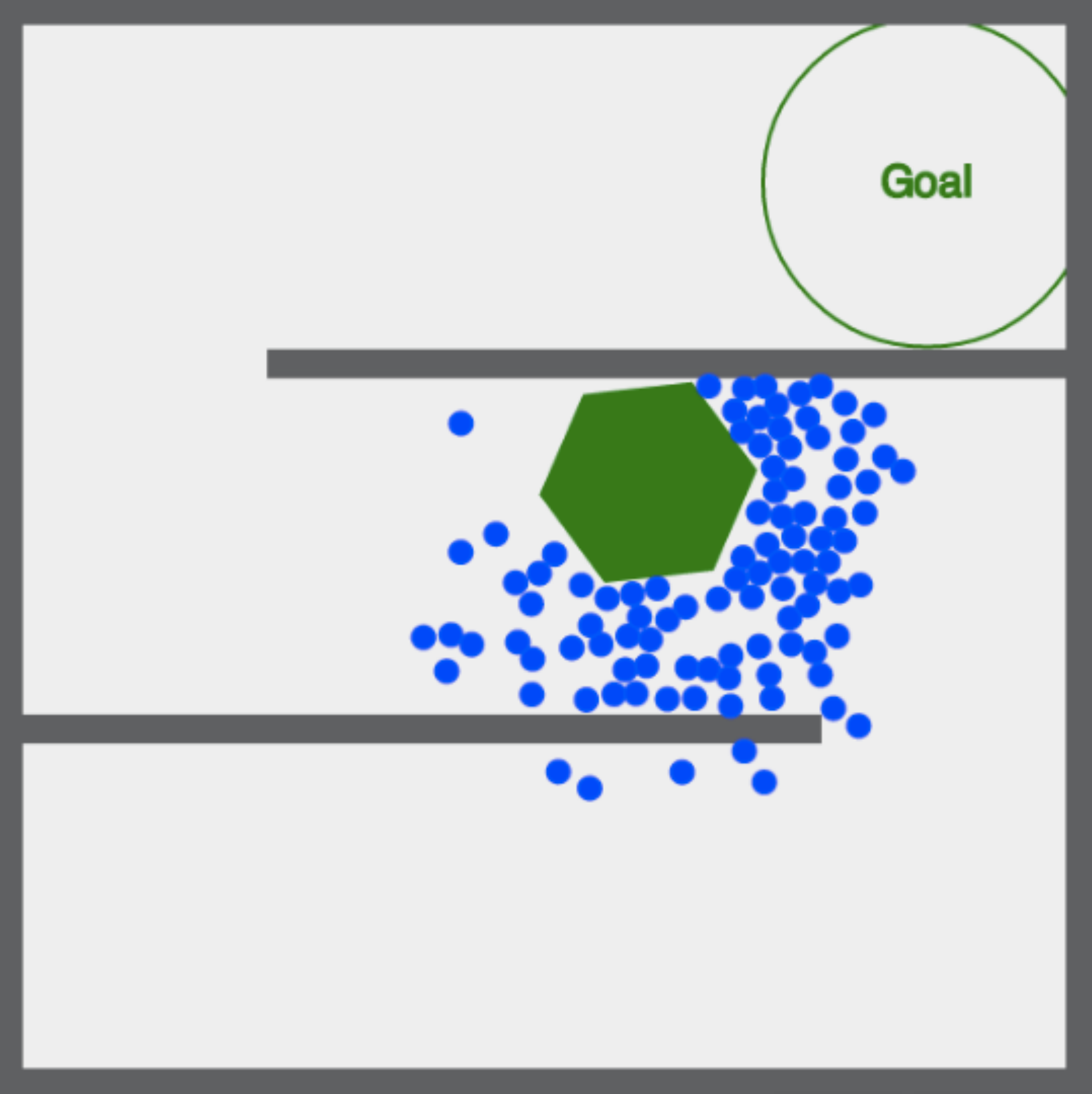}\end{overpic}
\begin{overpic}[width =\figwid]{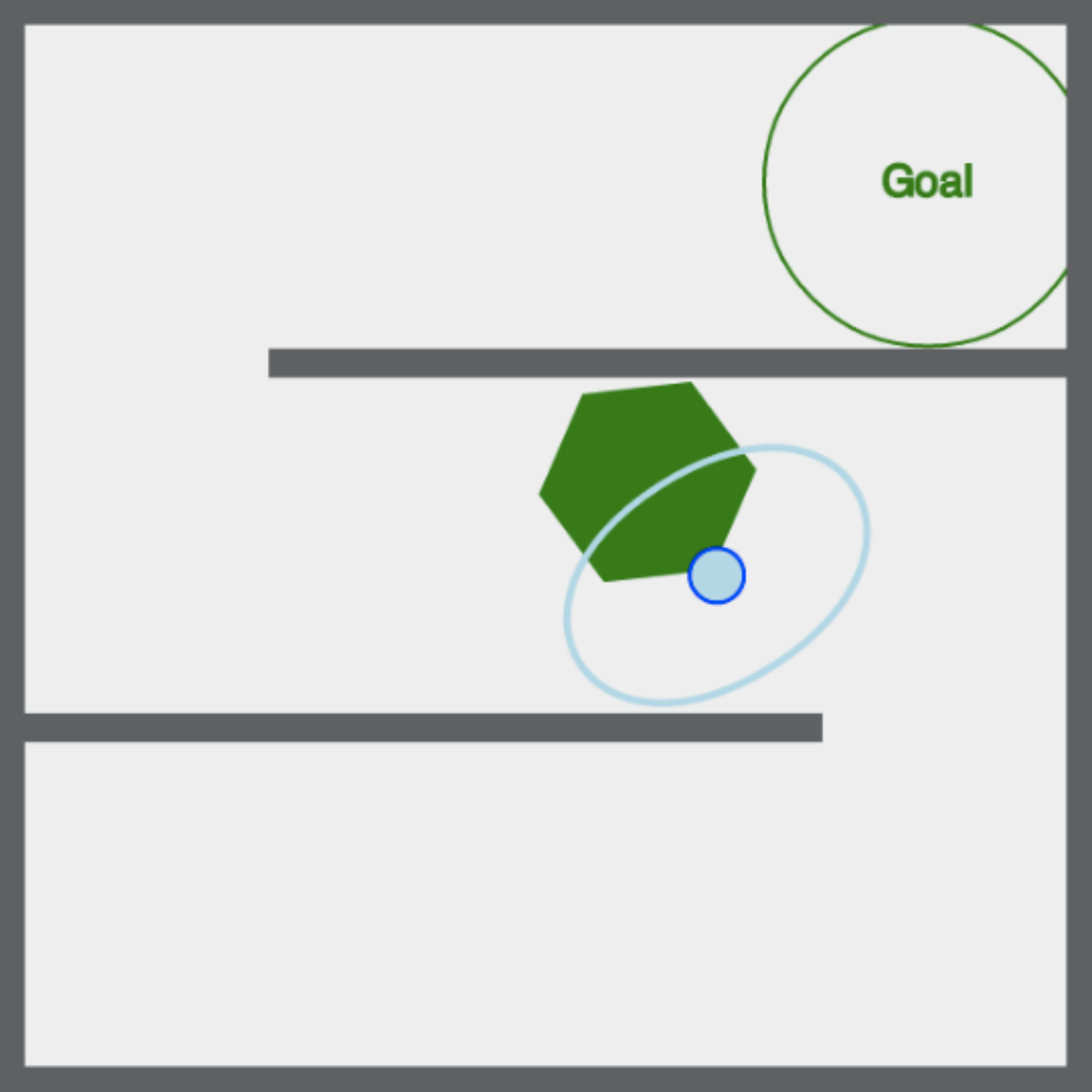}\end{overpic}}\\
\subfloat[][Vary Control]{\label{fig:VaryControl}
\begin{overpic}[width =\figwid]{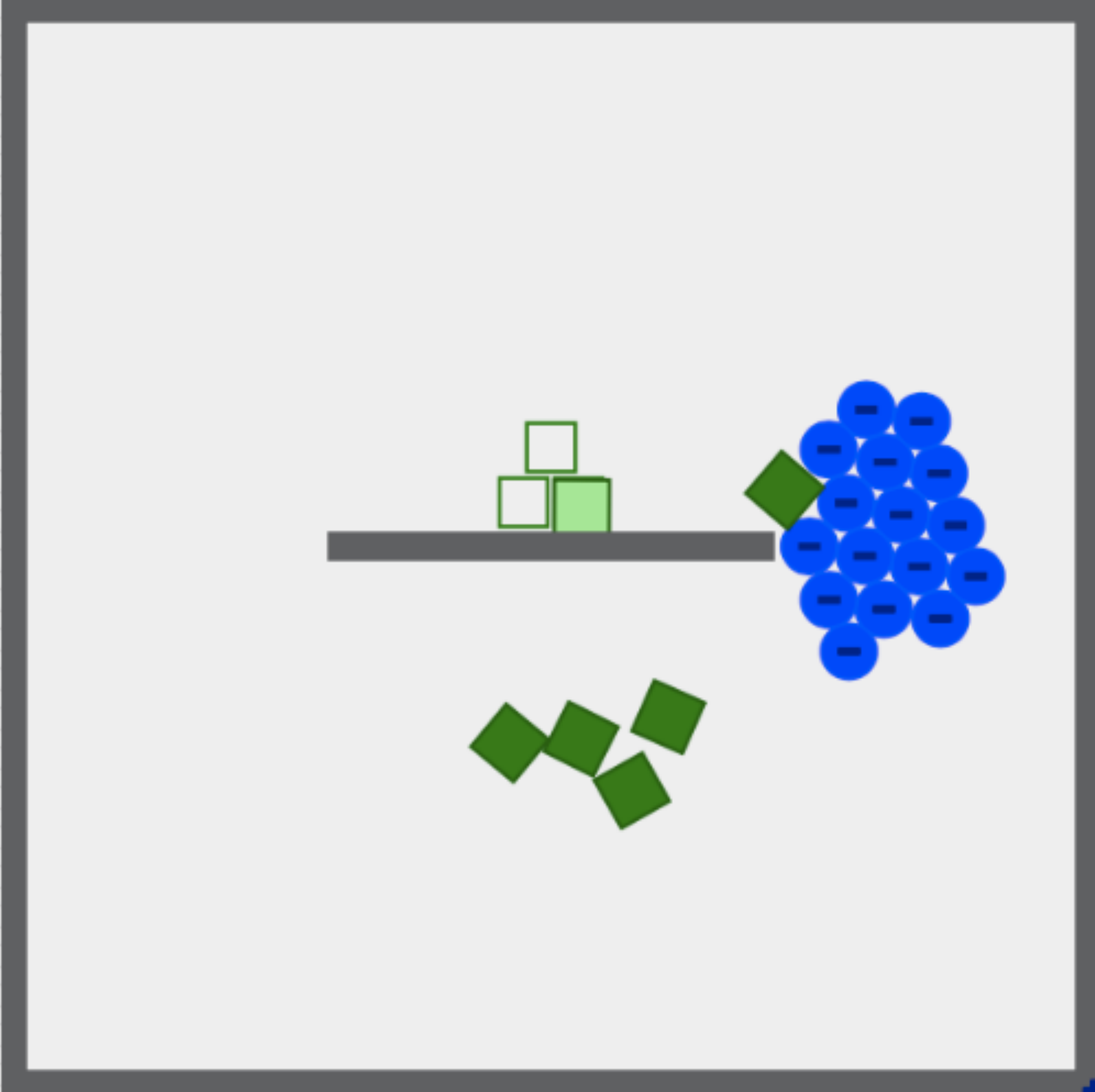}\end{overpic}}
\subfloat[][Vary Noise]{\label{fig:VaryNoise}
\begin{overpic}[width =\figwid]{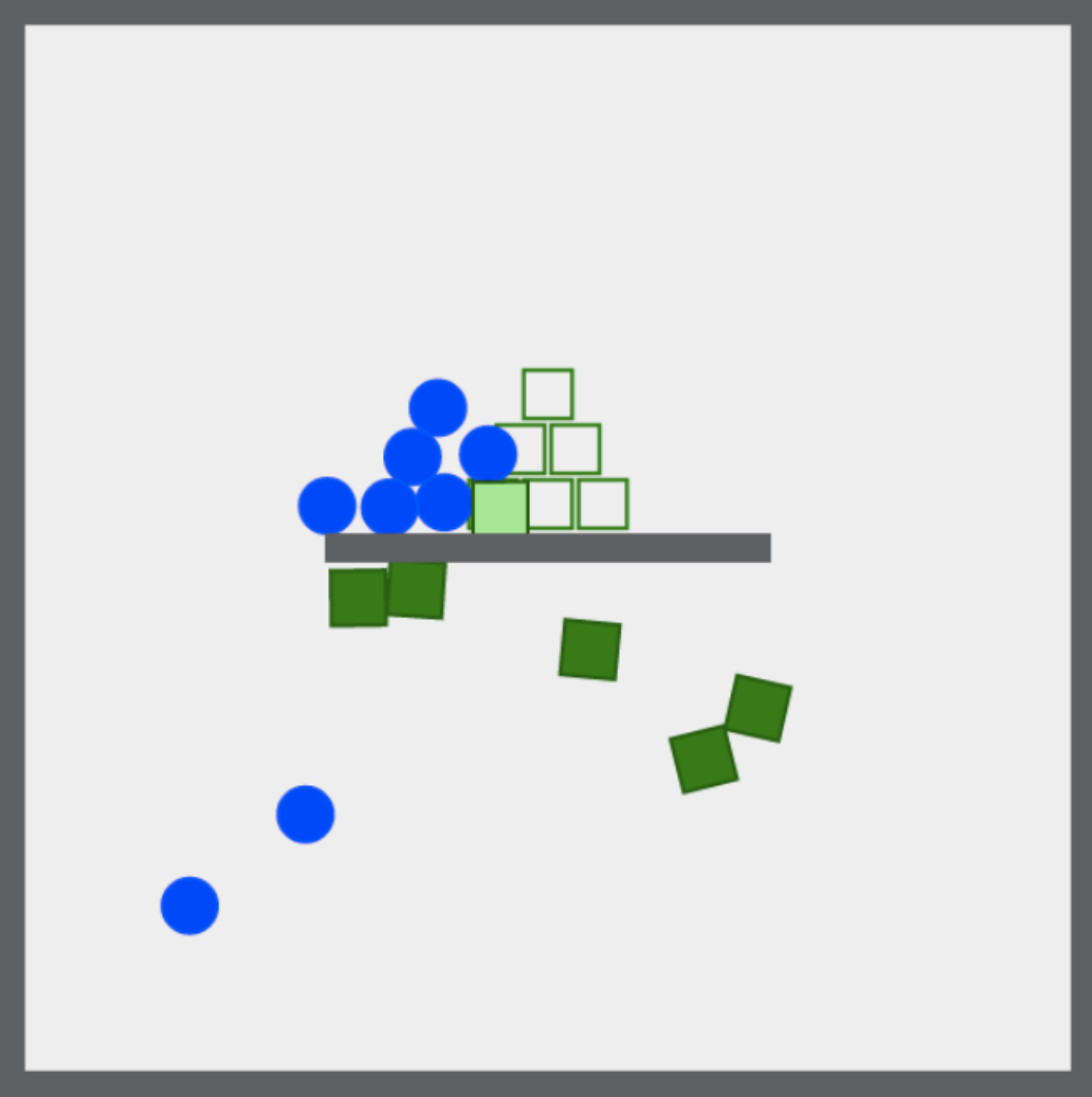}\end{overpic}}
\subfloat[][Control Position]{\label{fig:ControlPos}
\begin{overpic}[width =\figwid]{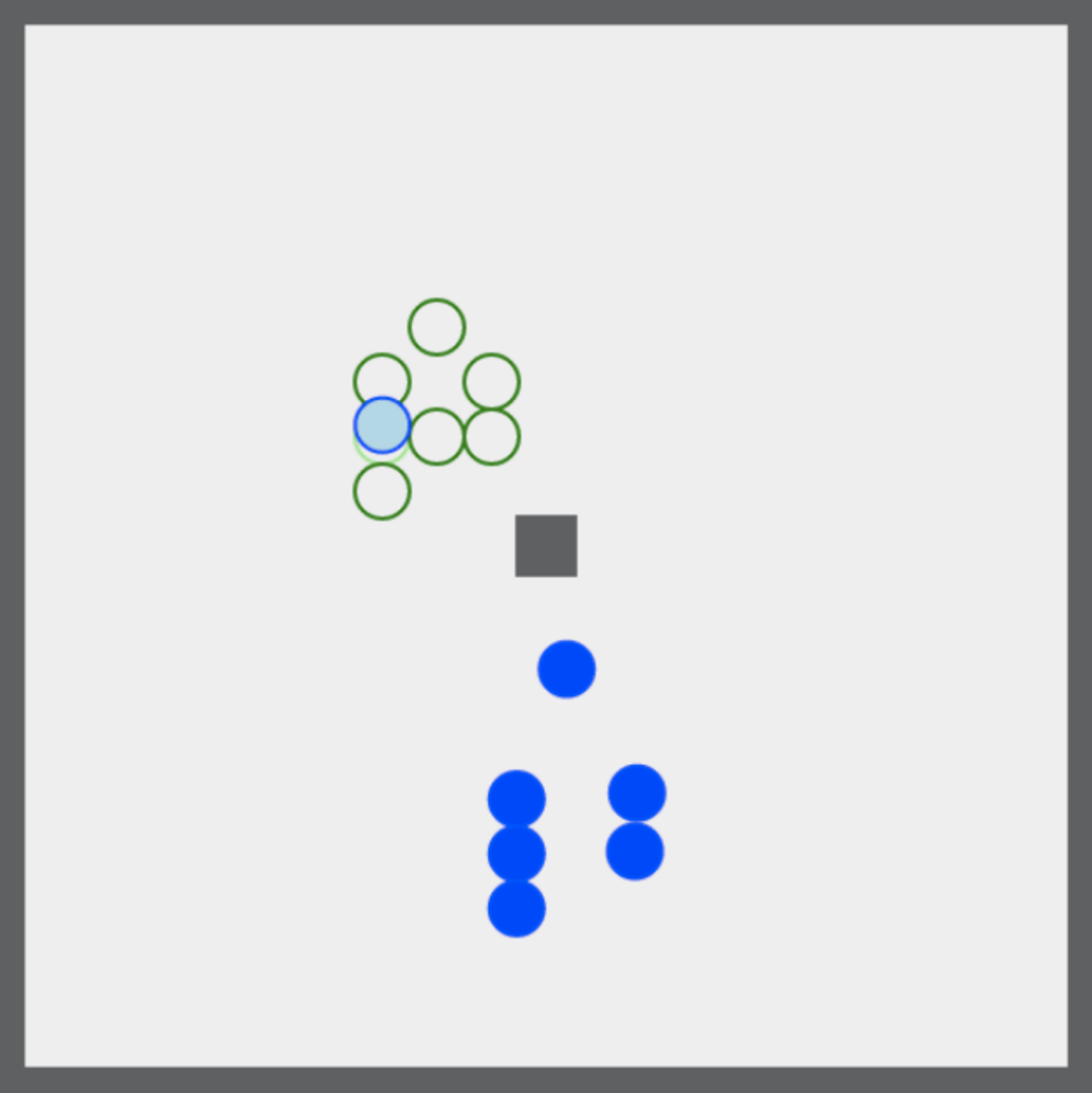}\end{overpic}}
\caption{\label{fig:5experiments}
Screenshots from our five online experiments controlling multi-robot systems with limited, global control.
\textbf{(a)} Varying the number of robots from 1-500
\textbf{(b)} Comparing 4 levels of visual feedback 
\textbf{(c)} Comparing 3 control architectures
\textbf{(d)} Varying noise from 0 to 200\% of control authority
\textbf{(e)} Controlling the position of 1 to 10 robots.
\href{http://youtu.be/HgNENj3hvEg}{See video overview at http://youtu.be/HgNENj3hvEg.}
\vspace{-2em}
}
\end{figure}

Our goal was to test several scenarios involving large-scale human-swarm interaction (HSI), and to do so with a statistically-significant sample size. Towards this end, we created \href{http://www.swarmcontrol.net/show_results}{SwarmControl.net}, an open-source online testing platform suitable for inexpensive deployment and data collection on a scale not yet seen in swarm robotics research. Screenshots from this platform are shown in Fig.~\ref{fig:5experiments}.  \href{https://github.com/crertel/swarmmanipulate.git}{All code}~\cite{Chris-Ertel2013}, \href{http://www.swarmcontrol.net/show_results}{and experimental results} are posted online.

Our experiments show that numerous simple robots responding to global control inputs are directly controllable by a human operator without special training, that the visual feedback of the swarm state should be very simple in order to increase task performance, and that humans perform swarm-object manipulation faster using attractive control schemes than repulsive control schemes.

 %function and implementation function

 Our paper is organized as follows.  After a discussion of related work in Section \ref{sec:RelatedWork}, we describe our experimental methods for an online human-user experiment in Section \ref{sec:expMethods}.  We report the results of our experiments in Section \ref{sec:expResults}, discuss the lessons learned in Section \ref{sec:discussion}, and end with concluding remarks in Section \ref{sec:conclusion}.

%%%%%%%%%%%%%%%
%%%%%%%%%%%%%%%

%%%%%%%%%%%%%%%%%%%%%%%%%%%%%%%%%%%%%%%%%%%%%%%%%%%%%%%%%%%
\section{Related Work}\label{sec:RelatedWork}
%%%%%%%%%%%%%%%%%%%%%%%%%%%%%%%%%%%%%%%%%%%%%%%%%%%%%%%%%%%

\subsection{Human-Swarm Interaction}
Olson and Wood studied human \emph{fanout}, the number of robots a single human user could control~\cite{Jr2004}.  %Studied as a key aspect of human-robot interaction, 
They postulated that the optimal number of robots was approximately  the autonomous time  divided by the interaction time required by each robot.  Their sample problem involved a multi-robot search task, where users could assign goals to robots.  Their user interaction studies with simulated planar robots  indicated a \emph{fanout plateau} of about 8 robots, after which there were diminishing returns.   They hypothesize that the location of this plateau is highly dependent on the underlying task, and our work indicated there are some tasks without plateaus. % (Lemmings is an example of this as well, for systems with autonomy
Their research investigated robots with 3 levels of autonomy.  We use robots without autonomy, corresponding with their first-level robots.

% Removed this -- 
%Chen, Barnes, and Harper-Sciarini published a review of supervisory control but emphasized high levels of autonomy, using the 10-level taxonomy given by \cite{Parasuraman2000}.  The direct control techniques this paper examines are the lowest level.

Squire, Trafton, and Parasuraman designed experiments showing that user-interface design had a high impact on the task effectiveness and the number of robots that could be controlled simultaneously in a multi-robot task \cite{Squire:2006:HCM:1121241.1121248}.

A number of user studies compare methods for controlling large swarms of simulated robots, for example \cite{bashyal2008human,kolling2012towards,de2012controllability}.  These studies provide insights but are limited by cost to small user studies; have a closed-source code base; and focus on controlling intelligent, programmable agents.  
For instance \cite{de2012controllability} was limited to a pool of 18 participants,  \cite{bashyal2008human} 5, and \cite{kolling2012towards} 32.
	Using an online testing environment, we conduct similar studies but with much larger sample sizes.

\subsection{Global-control of micro- and nanorobots}
Small robots have been constructed with physical heterogeneity so that they respond differently to a global, broadcast control signal.  Examples include \emph{scratch-drive microrobots}, actuated and controlled by a DC voltage signal from a substrate \cite{Donald2006,Donald2008};   magnetic structures  with different cross-sections that could be independently steered \cite{Floyd2011,Diller2013};   \emph{MagMite} microrobots with different resonant frequencies and a global magnetic field \cite{Frutiger2008}; and  magnetically controlled nanoscale helical screws constructed to stop movement at different cutoff frequencies of a global magnetic field
\cite{Tottori2012,Peyer2013}. 

Similarly, our previous work \cite{Becker2012,Becker2012k} focused on exploiting inhomogeneity between robots.  These control algorithms theoretically apply to any number of robots---even robotic continuums---but in practice process noise cancels the differentiating effects of inhomogeneity for more than tens of robots.  We desire control algorithms that extend to many thousands of robots.

  \subsection{Three challenges for massive manipulation}
 While it is now possible to create many micro- and nanorobots, there remain challenges in control, sensing, and computation. 
  
 \subsubsection{Control---global inputs}
 Many micro- and nanorobotic systems \cite{Tottori2012,Shirai2005,Chiang2011,Donald2006,Donald2008,Takahashi2006,Floyd2011,Diller2013,Frutiger2008,Peyer2013}
   rely on global inputs, where each robot receives an exact copy of the control signal.  Our experiments follow this global model.
   %Two reasonable questions are ``What tasks are possible with many robots, all under uniform control inputs?'' and ``What tasks are impossible with many robots, all under uniform control inputs?'' 
  
 \subsubsection{Sensing---large populations}
 Parallel control of $n$ differential-drive robots in a plane requires $3n$ state variables. Even holonomic robots require $2n$ state variables. Numerous methods exist for measuring this state in micro- and nanorobotics.  These solutions use computer vision systems to sense position and heading angle, with corresponding challenges of handling missed detections and image registration between detections and robots.  These challenges are increased at the nanoscale where sensing competes with control for communication bandwidth.   We examine control when the operator has access to partial feedback, including only the first and/or second moments of a population's position, or only the convex-hull containing the robots.
 
\subsubsection{Computation---calculating the control law}
In our previous work the controllers required at best a summation over all the robot states \cite{Becker2012k} and at worst a matrix inversion \cite{Becker2012}. 
These operations become intractable for large populations of robots. By focusing on \emph{human} control of large robot populations, we accentuate computational difficulties because the controllers are implemented by the unaided human operator. % We present an approach that, for many tasks, bypasses these problems due to large populations by allowing the user to command the entire population as a single unit.  For position control we cannot bypass this problem, but we provide an algorithm that scales linearly in the number of robots. 

\section{Experimental Methods}
\label{sec:expMethods}
%%%%%%%%%%%%%%%%%%%%%%%%%%%%%%%%%%%%%%%%%%%%%%%%%%%%%%%%%%%

% Experimental Methods
%%  Platform  
%%  Human Subjects
%%% recruited through social media
%%% IRB form  Protocol Number: 14-012E Protocol Title: Massive Manipulation: A n online user study on controlling large swarms of simple robot sApproval Date: 7/26/2013Expiration Date: 7/26/2014
%%% Costs  for experiment:  ??
%%% Instrumenting:
%%%% Google analytics, airbrake, etc.

%% wherein we describe our framework
\subsection{Framework}

We have developed a flexible testing framework for online human-swarm interaction studies. There are two halves to our framework: the server backend and the client-side (in-browser) frontend. The server backend is responsible for tabulating results, serving webpages containing the frontend code, and for issuing unique identifiers to each experiment participant. The in-browser frontend is responsible for running an experiment---that is to say, accepting user input, updating the state of the robot swarm, and ultimately evaluating task completion.

%% wherein we outline the process that a user takes to participate in an experiemnt
\subsubsection{Overview}

A participant visits the site, initiating a communication between their browser and our server. The web server generates a unique identifier for the participant and sends it along with the landing page to the participant---this identifier is stored as a browser cookie and will be sent along with all results the participant generates. The participant's browser prompts for confirmation of the terms-of-service and offers a menu of experiments.

Once the participant selects an experiment, their browser makes a new request to the server to load the experiment's webpage. The server sends scaffold HTML describing the layout of the page and a script block containing the experiment. The script runs the experiment and, upon a successful completion, posts the experiment data to the server. The participant is then given the option of playing again or trying a different experiment.

A participant may view all of the experimental data we have gathered; this information is available as either a webpage, a JSON file, or a comma-separated value file.

%% wherein we preach the good word of Ruby and describe what the backend actually does
\subsubsection{Backend}

The server backend is written in Ruby, using the Ruby-on-Rails (abbreviated \emph{Rails}) web development framework. Ruby is a dynamically-typed object-oriented scripting language with a strong emphasis on programmer ergonomics and metaprogramming support. It is well-suited for the creation of domain-specific languages for a variety of tasks, as exemplified by the Rails framework. Our backend serves assets (images, scripts, stylesheets, and so forth) to participants, selects the correct script to send to perform a particular experiment, and stores results.

Each result is a database record containing the experiment name, the participant identifier, the duration of the experiment (time to completion), the number of robots involved, the detailed mode information of the experiment, and the user agent string of the browser running the experiment (which identifies the type of browser used by the participant). Rails automates the process of creating the relevant database-object bindings, and thus we spent little time creating or modifying the result records, allowing us to rapidly adapt the server to our needs--for example, adding tracking of the user agent and experiment mode both took less than five minutes of work on the server side.

The experiment script file to be sent to the client is chosen with the uniform resource identifier (URI) for the experiment webpage; this done, the server will render the page requested by the participant and insert the script for the selected experiment. The Rails framework has a great deal of support for optimizing and compacting (\emph{minifying}) Javascript files.

%% wherein we decry the evils of javascript and explain why we use the browser
\subsubsection{Frontend}

The client frontend which runs in the participant's browser is written in Javascript, a dynamically-typed prototype-oriented scripting language with some functional programming support. We make heavy use of the Underscore framework (a functional programming toolkit for Javascript) as well as the Javascript port of \href{http://box2d.org/}{Box2D} (a popular 2D physics engine with good support for rigid-body dynamics and fixed-timestep simulation). Our frontend also includes helper libraries for drawing robots, handling user input, and drawing graphs.

Our framework uses a base task to represent the lifecycle of an experiment---{\bf  instantiation, simulation, evaluation}, and {\bf submission}. A particular experiment inherits from this prototype but overrides particular methods and adds its own variables for bookkeeping; this allows new or modified tasks to be created rapidly with minimal boilerplate code.

%\paragraph{Model}
Our robots are disc-shaped, non-holonomic, and confined to the 2D plane.  The control input $u$ consists of a single bounded force vector that is applied to each robot, $|u|\le u_{max}$.  We include a linear ramp for this force value that starts at zero and increases to the maximum value in one second; this allows participants to do fine control of the robots by tapping the arrow keys.
\begin{align}\label{eq:sysmodel}
\dot{x}_i = u_x,  \qquad  \dot{y}_i = u_y.
 \end{align}

During the {\bf instantiation} phase, an experiment sets up the web page elements with help text and other information, and creates the obstacles, robots, and workpieces that will be present during the experiment. It will also randomly select which mode to run in, if applicable.

The {\bf simulation} phase is the time at which all of the robots are moved according to user input and given a chance to interact with each other and the environment. The simulation phase then draws the current state of the experiment to the canvas of the webpage.

The {\bf evaluation} phase is when the experiment's completion criteria are applied to the current experiment state: are the robots in the goal zone, are the workpieces in the correct place, and so forth. If the criteria are not met, the experiment loops back into the simulation phase; if they are met, then the experiment proceeds to result submission.

The {\bf submission} phase is when the results of the experiment are combined with other user data, such as the browser user agent string, and submitted to the server for collection.

%As a means of encouraging user interaction, the results of other runs of the experiment are shown to the participant after submission along with merit badges displaying the number of experiments completed.

%% wherein we justify hours spent on facebook and hacker news  -- ha!  I'd like us to place links in the .tex for future reference
\subsection{Human subjects}
Because our study involved recording data from human subjects, it required IRB approval before we could legally save user data (IRB \#14-012E).
%% IRB form  Protocol Number: 14-012E Protocol Title: Massive Manipulation: An online user study on controlling large swarms of simple robot sApproval Date: 7/26/2013Expiration Date: 7/26/2014
%Rice Federal-Wide Assurance Number: 00003890

\begin{figure}
\centering
\begin{overpic}[width = .9\columnwidth]{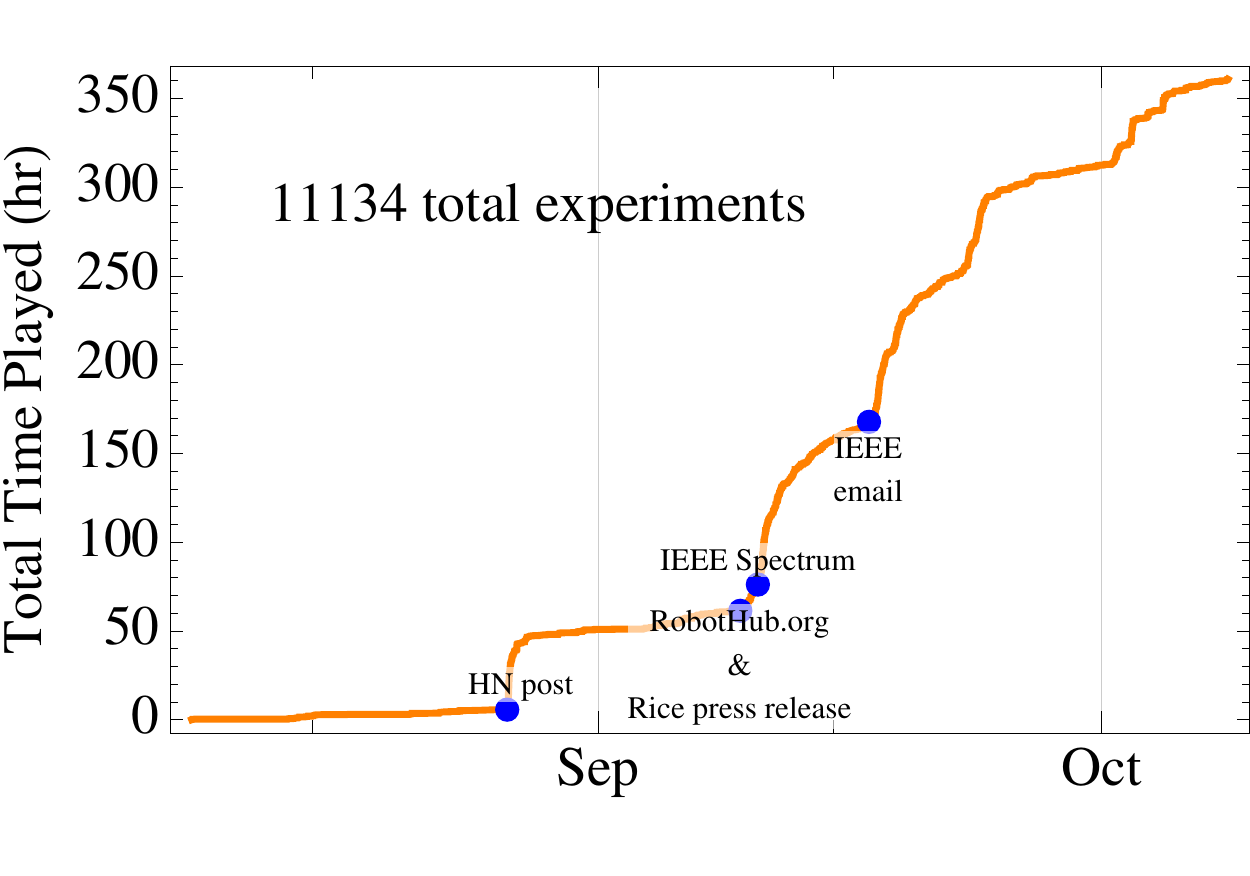}\end{overpic}
\vspace{-2em}
\caption{
\label{fig:timePlayed}
Cumulative time played for completed tests.
}
%\vspace{-2em}
\end{figure}

Subjects were recruited using a combination of social network effects and coordinated news posts. We asked our friends and colleagues to send links to our site to their friends via their preferred social networks, generally Twitter, Facebook, Google+, and through email. Additionally, we posted our site to several news aggregators in hopes that it would be seen and visited. Our first such posting was to \href{https://news.ycombinator.com/item?id=6277052}{Hacker News}, an aggregator run by the Y-Combinator accelerator company; this posting resulted in our first thousand trials. A second posting was made to Reddit, but did not seem to cause much traffic. A third posting was made to the  \href{http://robohub.org/researchers-use-single-joystick-to-control-swarm-of-rc-robots/}{Robohub.org} site. The traffic generated by these postings is shown in Fig.~\ref{fig:timePlayed}. 

Concurrently, we contacted our university's  \href{http://news.rice.edu/contact-us/}{\emph{News and Media Relations Team}}. They sent a writer and photographer to our lab, worked with us to draft a \href{http://news.rice.edu/2013/09/09/a-swarm-on-every-desktop-robotics-experts-learn-from-public/ }{press release}, and publicized with news outlets and alumni. Most universities have a media team, and this is a valuable no-cost resource to gain publicity.
 
%% wherein we show just how cheap we are 
\subsection{Experimental costs}

We've spent approximately one hundred dollars USD provisioning and running this experiment. Hosting is provided by \href{Heroku.com}{Heroku}, using a single web instance costing around \$40/month, with additional monitoring services bringing that up to \$50/month. In the event of increased demand/participant traffic, we can provision another server to take up the load.
We purchased our domain name from \href{Namecheap.com}{Namecheap.com} for \$11.66 a year, giving our site a short, easy to pronounce handle.
%SwarmControl.net %16 char
%SwarmControl.herokuapp.com %26 char

%As mobile traffic becomes more prominent and overtakes the volume of internet access from desktop computers & laptops, shorter domain names will become increasingly more important and more desirable.  The likelihood of a typo on a mobile device is increased and that “liability” can be hedged by having a shorter domain name which requires less typing and fewer characters to enter. - See more at: http://www.mediaoptions.com/domain-names/short-is-sweet-the-value-of-short-domain-names.html#sthash.AJNqlXhB.dpuf

Given the large number of experiment sessions run (over 11,000 at the time of this writing), we see a per-experiment cost of less than three cents.

%% wherein we describe how we monitor the progress of the experimental setup
\subsection{Instrumentation}

When conducting an online experiment, it is very helpful to gather data about both the experiment infrastructure and the participants. For the backend, we use a service called Airbrake to monitor the `health' of the Rails server, getting emails in the events of any errors occurring or suspicious activity. We also use another service called New Relic to provide monitoring and analytics on the server traffic, giving coarse statistics about site visitation, page load time, and other indicators of how our backend is performing.

For the frontend, we use Google Analytics to track user behavior. This tool allows us to see country of origin for users, time spent on the site, relative percent of people who look past the landing page (`bounce rate'), and user agent information (type of browser, type of device, etc.).

%%%%%%%%%%%%%%%
%%%%%%%%%%%%%%%

%%%%%%%%%%%%%%%%%%%%%%%%%%%%%%%%%%%%%%%%%%%%%%%%%%%%%%%%%%%
\section{Results}\label{sec:expResults}
%%%%%%%%%%%%%%%%%%%%%%%%%%%%%%%%%%%%%%%%%%%%%%%%%%%%%%%%%%%

\begin{figure}
\begin{overpic}[width = 0.48\columnwidth]{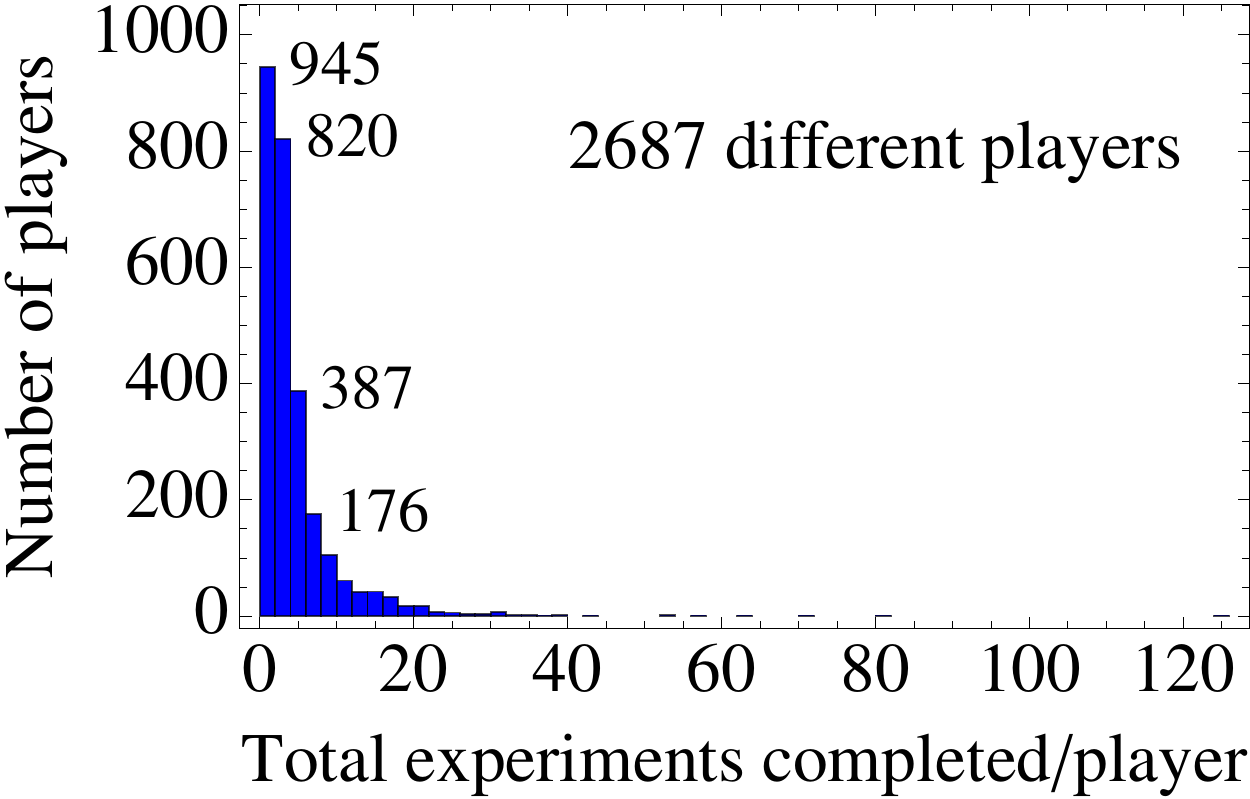}\end{overpic}
\begin{overpic}[width = 0.48\columnwidth]{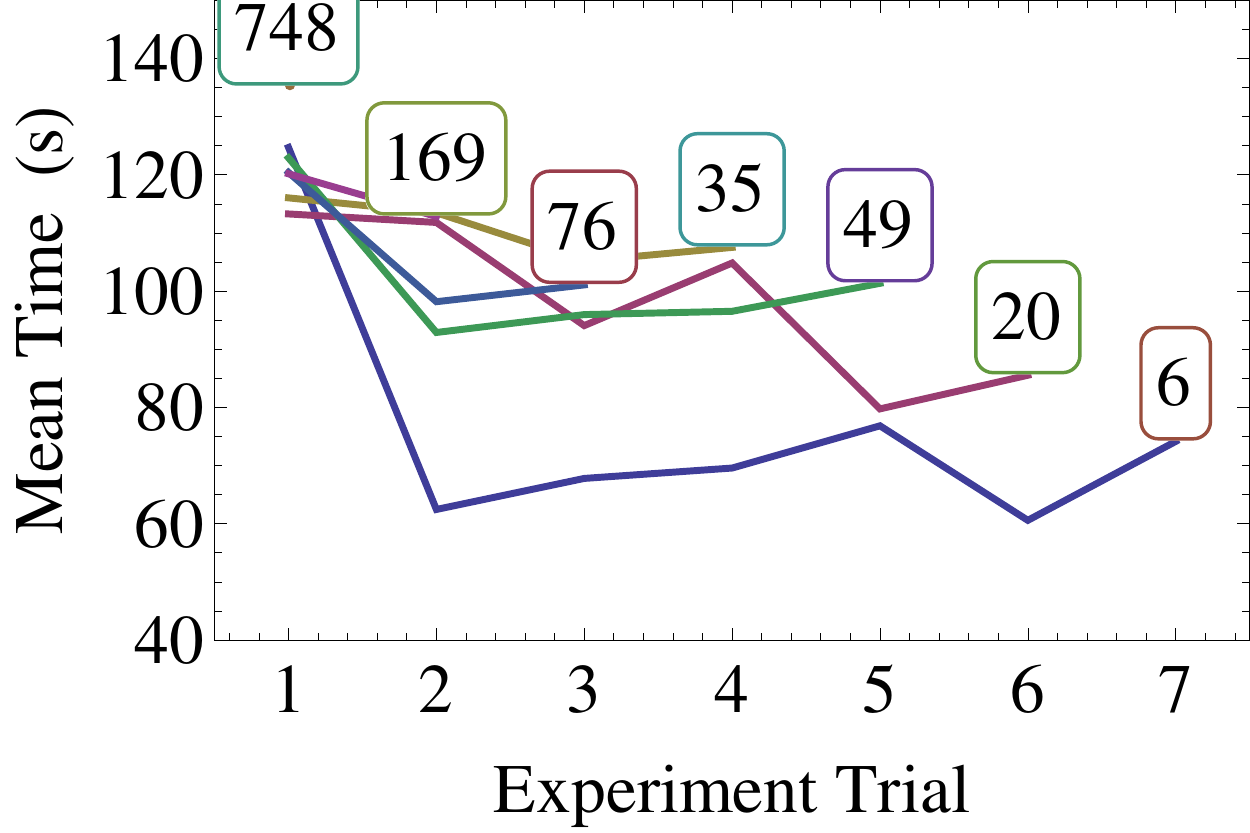}\end{overpic}
\caption{
\label{fig:Learning}
(Left) The total number of games played per player drops off exponentially. (Right) We are able to show that players skill improves as they retry tests using data from \emph{Varying Number}
}
\end{figure}

We designed five experiments to investigate human control of large robotic swarms for manipulation tasks.  Screenshots of each experiment are shown in Fig.~\ref{fig:5experiments}.  Each experiment examined the effects of varying a single parameter: population of robots for manipulation, four levels of visual feedback, three control architectures, different levels of Brownian noise, and position control with 1 to 10 robots. The users could choose which experiment to try, and then our architecture randomly assigned a particular parameter value for each trial.  We recorded the completion time and the participant ID for each successful trial.  As Fig.~\ref{fig:Learning} shows, one-third of all participants played only a single game.  Still, many played multiple games, and their decreasing completion times demonstrates their skills improved.

% RESULTS
%% For each task, list the hypothesis, then give plot

%\begin{figure}
%\centering
%\begin{overpic}[width = 0.5\columnwidth]{DesignSpace.pdf}\end{overpic}
%\caption{
%\label{fig:DesignSpace} 
%There is a multi-dimensional design space for controlling robot swarms.  In this paper we provide a framework for exploring this design space. To demonstrate the capabilities of this framework, in Section \ref{sec:experiments} we describe experiments that test task completion time as a function of the number of robots, the quality of the state feedback, and the control architecture.}
%\end{figure}

\subsection{Varying Number}
\begin{figure}
\centering
\begin{overpic}[width = \columnwidth]{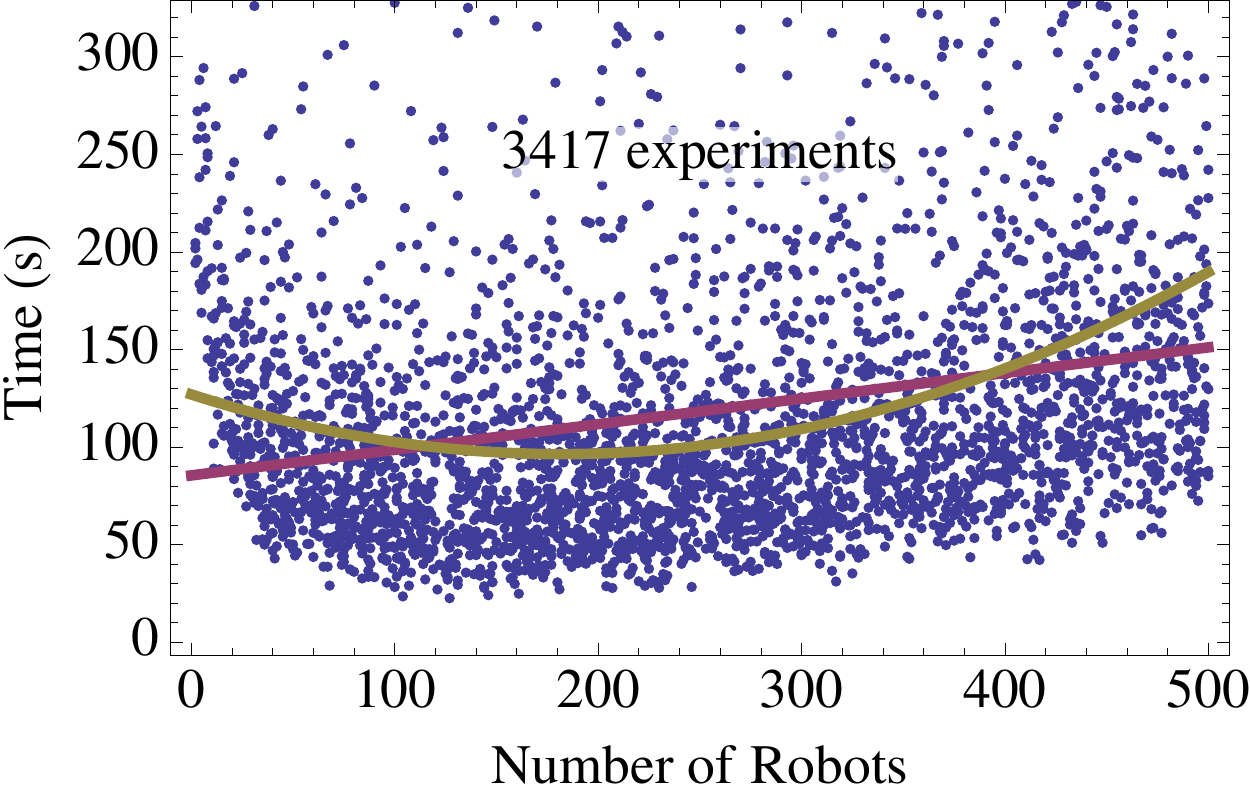}\end{overpic}
\vspace{-1em}
\caption{\label{fig:ResVaryNu}Data from \emph{Varying Number} using robots to push an object through a maze to a goal location.  The data indicates that this task has an optimal number of robots, perhaps due to the relative sizes of the robots, obstacles, and object. Best-fit linear and quadratic lines are overlaid for comparison. 
%\vspace{-1em}
}
\end{figure}

Transport of goods and materials between points is at the heart of all engineering and construction in real-world systems. This experiment varied from 1 to 500 the population of robots used to transport an object. We kept the total area, maximum robot speed, and total net force the swarm could produce constant. The robots pushed a large hexagonal object through an `S'-shaped maze. Our hypothesis was that participants would complete the task faster with more robots. The results, shown in Fig.~\ref{fig:ResVaryNu}, do not support our hypothesis, indicating rather that there is a local minima around 130 robots.

\subsection{Varying Control}
Ultimately, we want to use swarms of robots to build things. This experiment compared different control architectures modeled after real-world devices.

We compared attractive and repulsive control with the global control used for the other experiments. The attractive and repulsive controllers were loosely modeled after scanning tunneling microscopes (STM), but also apply to magnetic manipulation~\cite{Khalil2013} and biological models~\cite{goodrich2012types}. STMs can be used to arrange atoms and make small assemblies~\cite{avouris1995manipulation}. An STM tip is charged with electrical potential, and used to repel like-charged or to attract differently-charged molecules. In contrast, the global controller uses a uniform field (perhaps formed by parallel lines of differently-charged conductors) to pull molecules in the same direction.
The experiment challenged players to assemble a three-block pyramid with a swarm of 16 robots.

%\todo{describe controller model here, use an equation}

\begin{figure}
\centering
\begin{overpic}[width = \columnwidth]{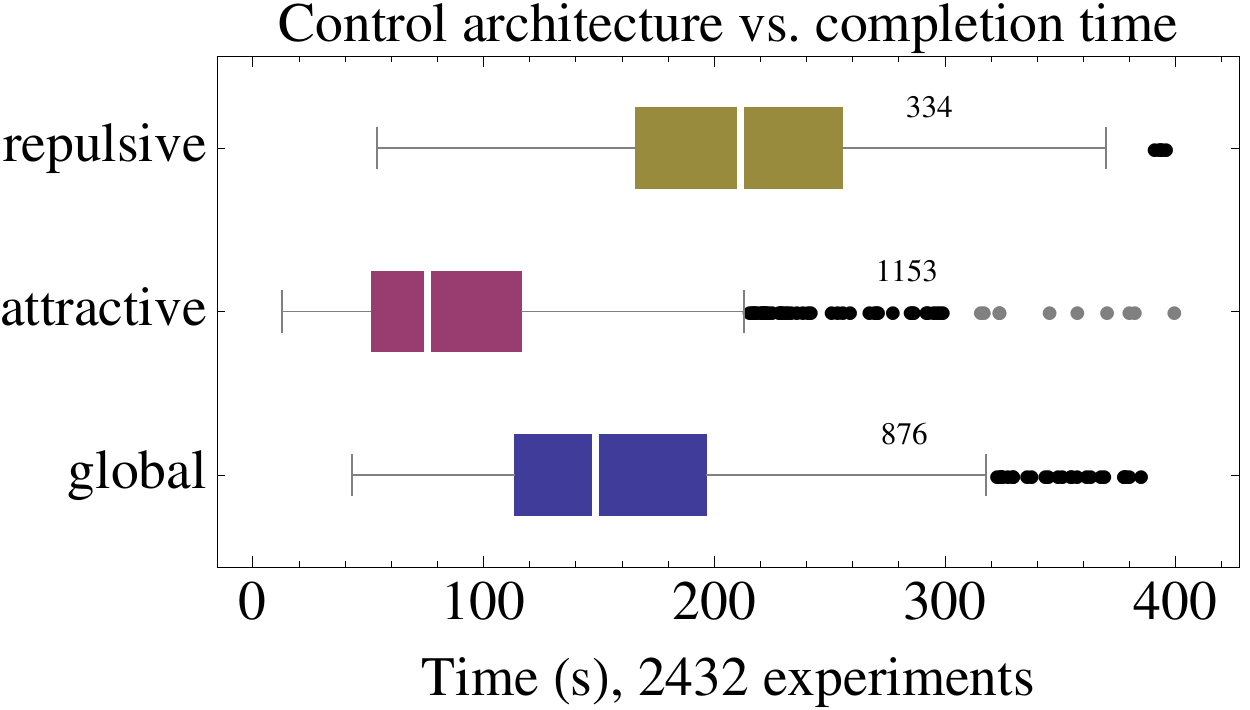}\end{overpic}
\vspace{-1em}
\caption{\label{fig:ResVaryControl}Attractive control resulted in the shortest completion time and repulsive the longest for building a three-block pyramid.
% \vspace{-1em}
}
\end{figure}

The results were conclusive, as shown in Fig.~\ref{fig:ResVaryControl}: attractive control was the fastest, followed by global control, with repulsive control a distant last.  The median time using repulsive control was four times longer than with attractive control.

%%\begin{figure}
%%\begin{overpic}[width = 0.39\columnwidth]{attractiveForce.pdf}\end{overpic}
%%\begin{overpic}[width = 0.6\columnwidth]{understandingForces.pdf}\end{overpic}
%%\caption{
%%\label{fig:attractiveForce}
%%Attraction and repulsion with a point source distance $h$ above the plane and $r$ from the robot. 
%%}
%%\end{figure}

\subsection{Varying Visualization}

\begin{figure}[b!]
\renewcommand{\figwid}{0.24\columnwidth}
\begin{overpic}[width =\figwid]{VaryVisFS.pdf}\put(20,15){Full-state}\end{overpic}
\begin{overpic}[width =\figwid]{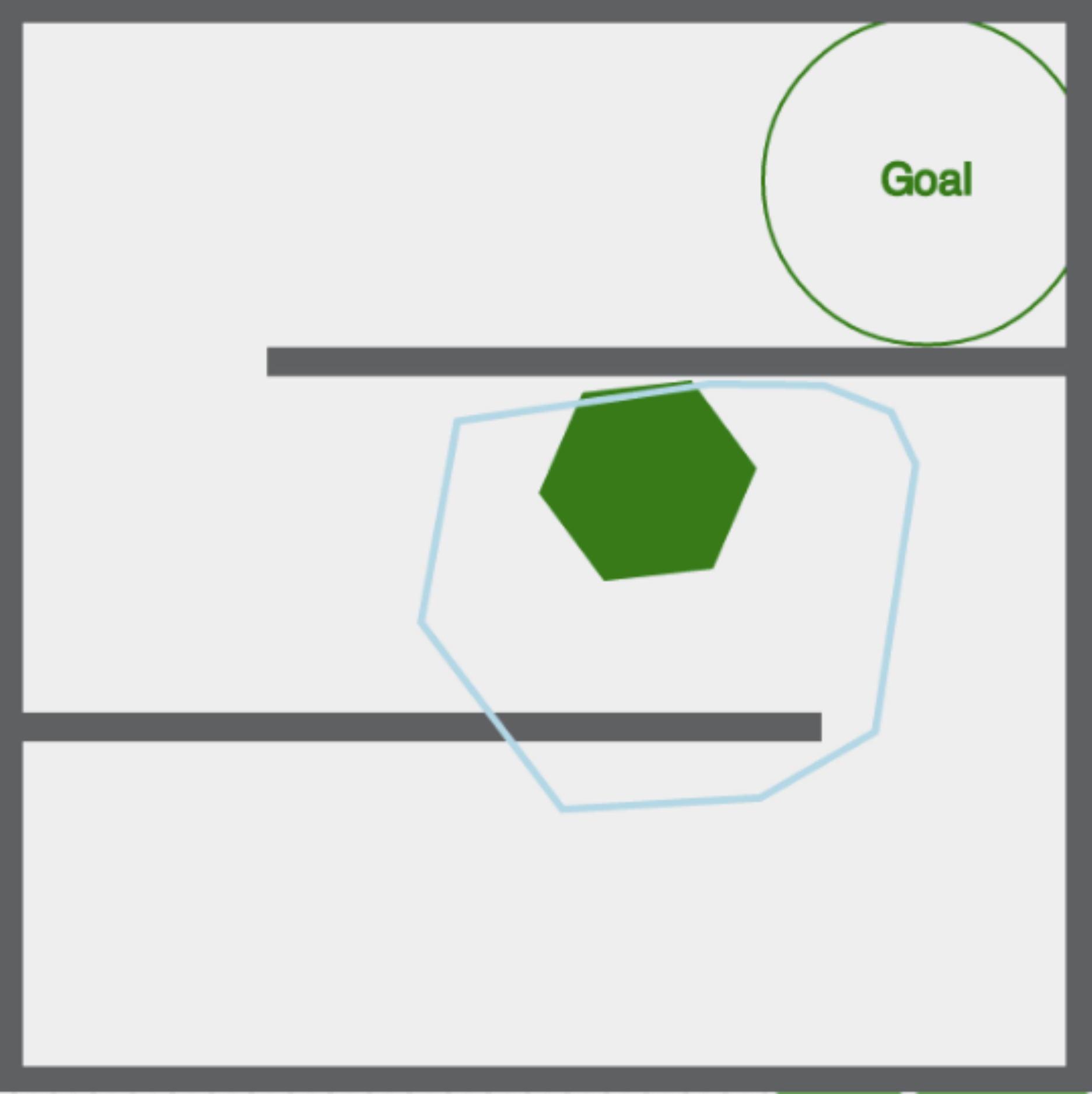}\put(10,15){Convex-hull}\end{overpic}
\begin{overpic}[width =\figwid]{VaryVisMV.pdf}\put(10,15){Mean + var}\end{overpic}
\begin{overpic}[width =\figwid]{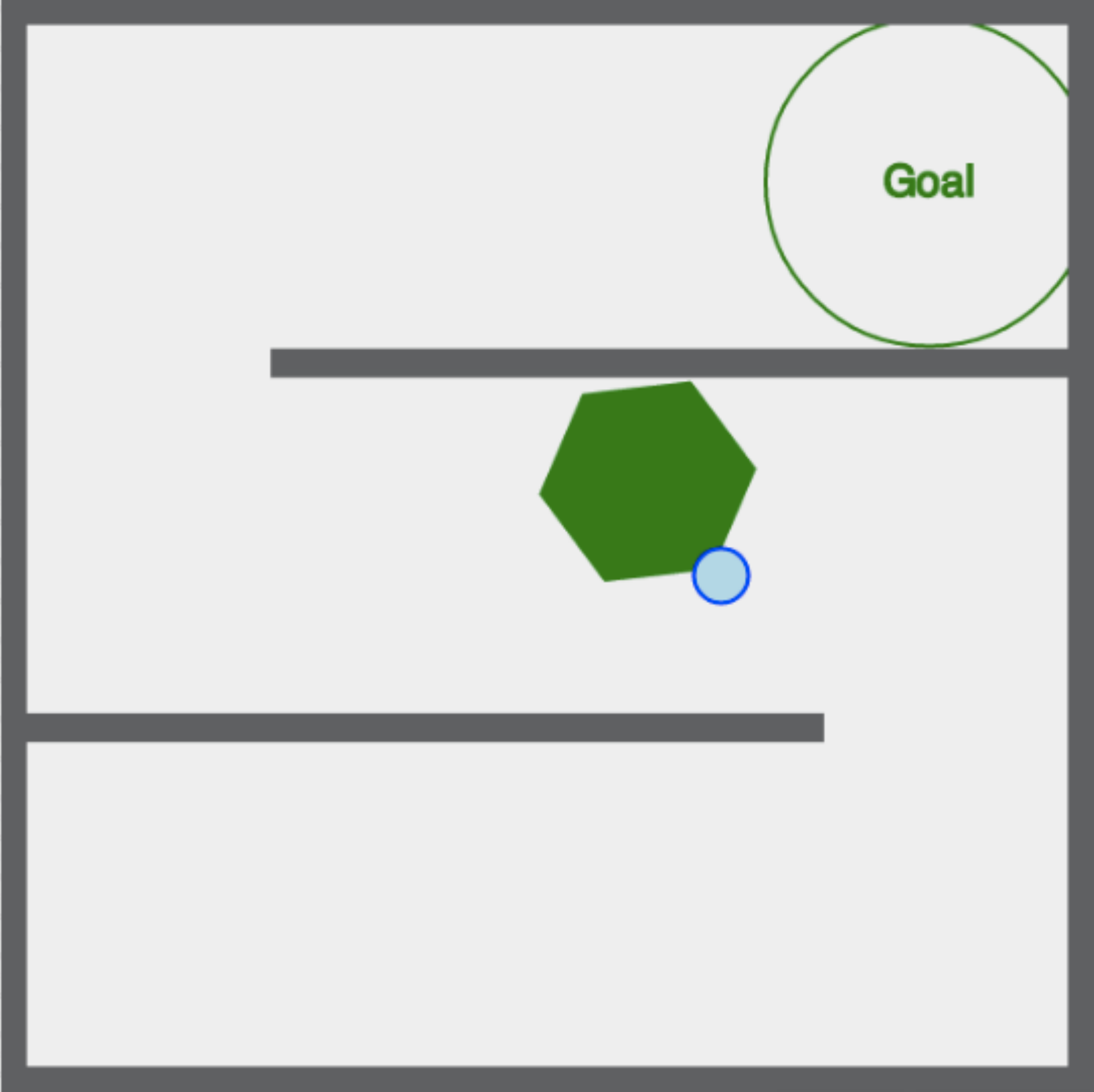}\put(30,15){Mean}\end{overpic}
\vspace{-2em}
\caption{\label{fig:Visualization}Screenshots from task \emph{Vary Visualization}. This experiment challenges players to quickly steer 100 robots (blue discs) to push an object (green hexagon) into a goal region. We record the completion time and other statistics.
%\vspace{-1em}
}
\end{figure}

Sensing is expensive, especially on the nanoscale. To see nanocars~\cite{Chiang2011}, scientists fasten molecules that fluoresce light when activated by a strong light source. Unfortunately, multiple exposures can destroy these molecules, a process called \emph{photobleaching}. Photobleaching can be minimized by lowering the excitation light intensity, but this increases the probability of missed detections~\cite{Cazes2001}.  This experiment explores manipulation with varying amounts of sensing information: {\bf full-state} sensing provides the most information by showing the position of all robots; {\bf convex-hull} draws a convex hull around the outermost robots; {\bf mean} provides the average position of the population; and {\bf mean + variance} adds a confidence ellipse. Fig.~\ref{fig:Visualization} shows screenshots of the same robot swarm with each type of visual feedback. Full-state requires $2n$ data points for $n$ robots. Convex-hull requires at worst $2n$, but usually a smaller number.  Mean requires two, and variance three, data points.  Mean and mean + variance are convenient even with millions of robots. Our hypothesis predicted a steady decay in performance as the amount of visual feedback decreased.

% Additionally, as population characteristics, they are available even if only a percentage of the robots are detected each control cycle.
%Photobleaching: http://www.piercenet.com/browse.cfm?fldID=4DD9D52E-5056-8A76-4E6E-E217FAD0D86B
%
%Photobleaching is caused by the irreversible destruction of fluorophores due to either the prolonged exposure to the excitation source or exposure to high-intensity excitation light. Photobleaching can be minimized or avoided by exposing the fluor(s) to the lowest possible level of excitation light intensity for the shortest length of time that still yields the best signal detection; this requires optimization of the detection method using high sensitivity CCD cameras, high numerical aperture objective and/or the widest bandpass emission filter(s) available. Other approaches include using fluorophores that are more photostable than traditional fluorophores and/or using antifade reagents to protect the fluor(s) against photobleaching. Steps to avoid photobleaching are not feasible for all detection methods and should be optimized for each method used. For example, antifade reagents are toxic to live cells, and therefore they can only be used with fixed cells or tissue. Furthermore, some detection methods, such as flow cytometry, normally do not require steps to avoid photobleaching because of the extremely short exposure time of the fluorophore to the excitation source.

\begin{figure}
\centering
\begin{overpic}[width = \columnwidth]{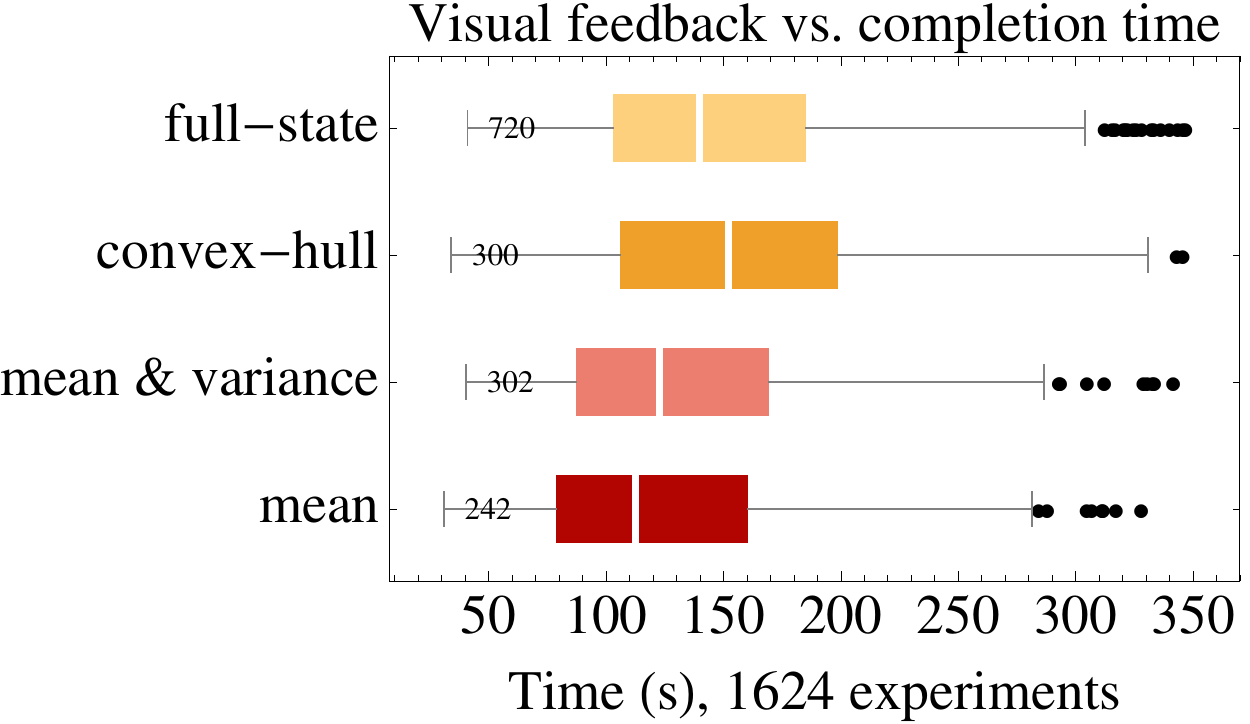}\end{overpic}
\vspace{-2em}
\caption{\label{fig:ResVaryVis} Completion-time results for the four levels of visual feedback shown in Fig.~\ref{fig:Visualization}. Surprisingly, players perform better with limited feedback--subjects with only the mean + variance  outperformed all others.
%\vspace{-2em}
}
\end{figure}

To our surprise, our experiment indicates the opposite: players  with just the mean completed the task faster than those with full-state feedback.  As Fig.~\ref{fig:ResVaryVis} shows, the levels of feedback arranged by increasing completion time are [mean + variance, mean, full-state, convex-hull].  Anecdotal evidence from beta-testers who played the game suggests that tracking 100 robots is overwhelming---similar to schooling phenomenons that confuse predators---while working with just the mean + variance is like using a ``spongy'' manipulator. Our beta-testers found convex-hull feedback confusing and irritating.  A single robot left behind an obstacle will stretch the entire hull, obscuring the majority of the swarm.
%obscuring what the rest of the swarm is doing.   

\subsection{Varying Noise}
Real-world microrobots and nanorobots are affected by turbulence caused by random collisions with molecules. The effect of these collisions is called Brownian motion.

This experiment varied the strength of these disturbances to study how noise affects human control of large swarms. Noise was applied at every timestep as follows:
\begin{align*}
\dot{x}_i &= u_x + m_i\cos(\psi_i)\\
 \dot{y}_i &= u_y + m_i\sin(\psi_i).
 \end{align*}
Here $m_i,\psi_i$ are uniformly IID, with $m_i\in[0,M]$ and $\psi_i\in[0,2\pi]$, where $M$ is a constant for each trial ranging from 0 to 200\% of the maximum control power ($u_{max}$).
 
\begin{figure}
\centering
\begin{overpic}[width = \columnwidth]{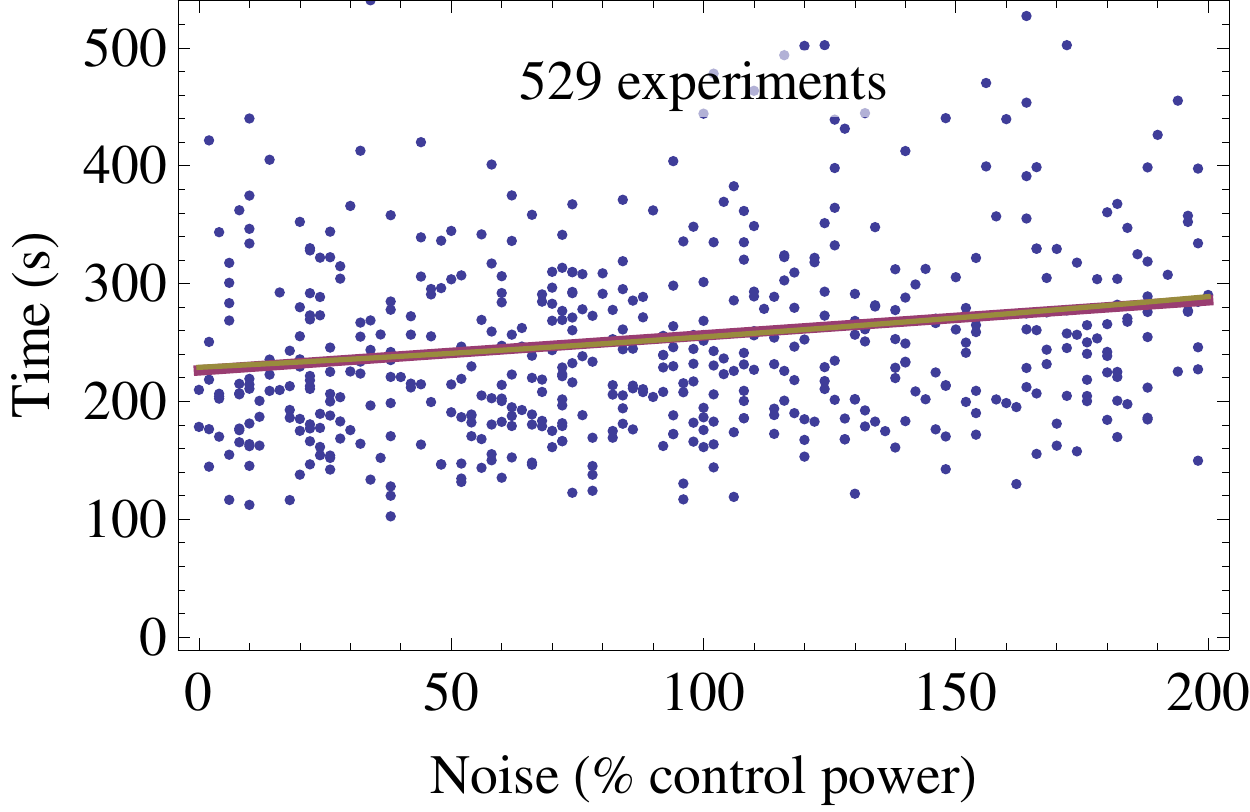}\end{overpic}
\vspace{-2em}
\caption{\label{fig:ResVaryNoise} Varying the noise from 0 to 200\% of the maximum control input resulted in only a small increase in completion time.
%\vspace{-2em}
}
\end{figure}

We hypothesized 200\% noise was the largest a human could be expected to control---at 200\% noise, the robots move erratically.  Disproving our hypothesis, the results in Fig.~\ref{fig:ResVaryNoise} show only a 40\% increase in completion time for the maximum noise.

\subsection{Position Control}
\begin{figure}
\centering
\begin{overpic}[width = \columnwidth]{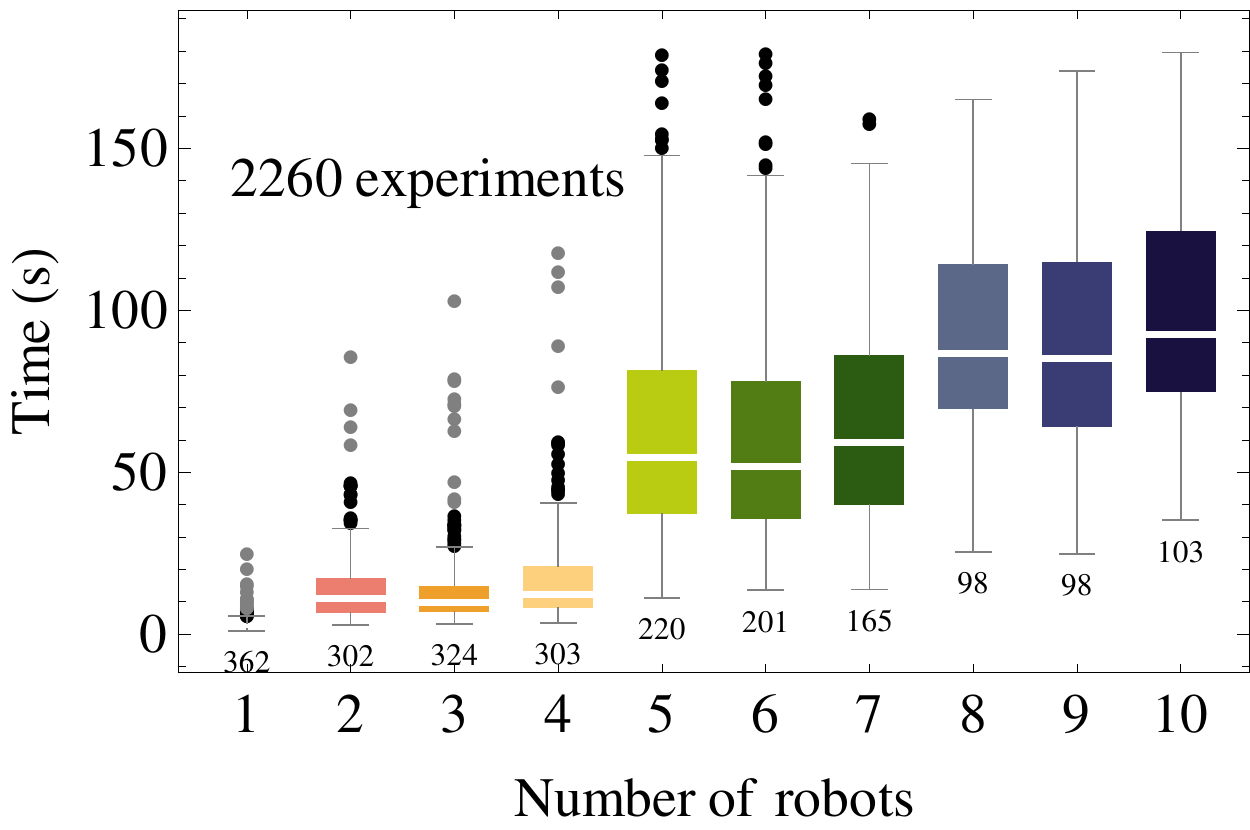}\end{overpic}
\vspace{-2em}
\caption{\label{fig:ResPositioning}Increasing the number of robots resulted in longer completion times.  For more than 4 robots the goal pattern contained a void, which may have caused the longer completion times.
%\vspace{-2em}
}
\end{figure}

This experiment examined how completion time scales with the number of robots $n$. Using a single square obstacle, users arranged $n\in[1,10]$ robots into a specified goal pattern.  The goal pattern formed a block `A' with 10 robots, and lesser numbers of robots used a subset of these goal positions. Our hypothesis was that completion time would increase linearly with the number of robots, as with our position control algorithm in \cite{Becker2013b}.  Our results roughly corroborate this, as shown in Fig.~\ref{fig:ResPositioning}.  Though the number of robots presented to game players is uniformly distributed, larger $n$ are more difficult, and the number of successful experiments drops steadily as $n$ increases.

Note there is a bifurcation between $n$=4 and $n$=5 robots. For $n\in[1,4]$ the goal patterns are not hollow, but starting at $n$=5 they are.  A better experiment design would randomly place the goal positions.  Initially we tried this, but our beta-testers strongly disliked trying to arrange robots in random patterns.

%%%%%%%%%%%%%%%
%%%%%%%%%%%%%%%
\section{Discussion}\label{sec:discussion}

During beta-testing and live deployment of SwarmControl.net we learned many things which will inform future experimentation. These lessons are directly applicable to designing better experiments, getting more results, and reaching more users.

\subsection{Lessons learned: Beta-testing}

While waiting on our IRB approval, we tested our experiment suite.  This gave us an opportunity to refine our software and to fix many flaws before they would have affected a larger audience; none of this data was used in our analysis.

\subsubsection{Be accessible}

Our original color scheme made heavy use of bright red, green, and blue colors. We tested for colorblind accessibility, and changed our color scheme. \href{http://colorfilter.wickline.org/}{http://colorfilter.wickline.org/} is an online service which displays a website as it appears to a colorblind user.

\subsubsection{Have simple instructions}

\begin{figure}
\begin{overpic}[width = \columnwidth]{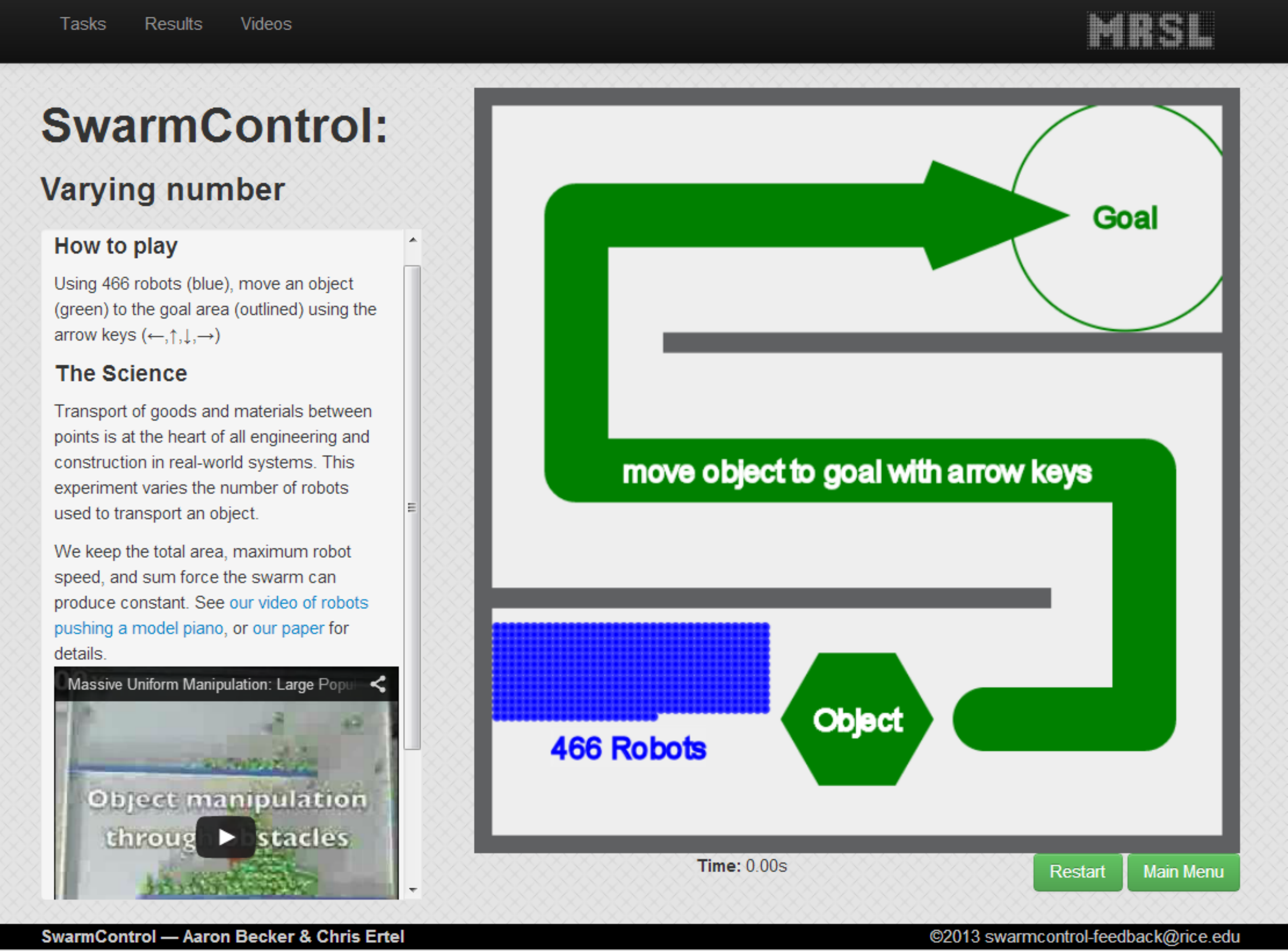}\end{overpic}
\vspace{-2em}
\caption{\label{fig:ScreenShotBeforePlay}Screenshot showing the \emph{Varying Number} experiment before gameplay.  At left are instructions and educational material, including a video. At right is the experiment canvas. Large instructions and bold arrows demonstrate the desired task---participants do not bother to read small print. 
\vspace{-1em}
}
\end{figure}

Beta-testers never looked at the instructions in the side-panel. To fix this, we overlaid simple, mostly pictorial instructions on the experiment canvas before play, as shown in Fig.~\ref{fig:ScreenShotBeforePlay}.

\subsubsection{Make the structure obvious}

Beta-testers desired a more structured experience; initially we had only shown screenshots of the five experiments and each experiment had a replay button. Participants told us that they had no idea when they had finished, how they had performed, or why the experiment mattered.

\begin{figure}
\begin{overpic}[width = \columnwidth]{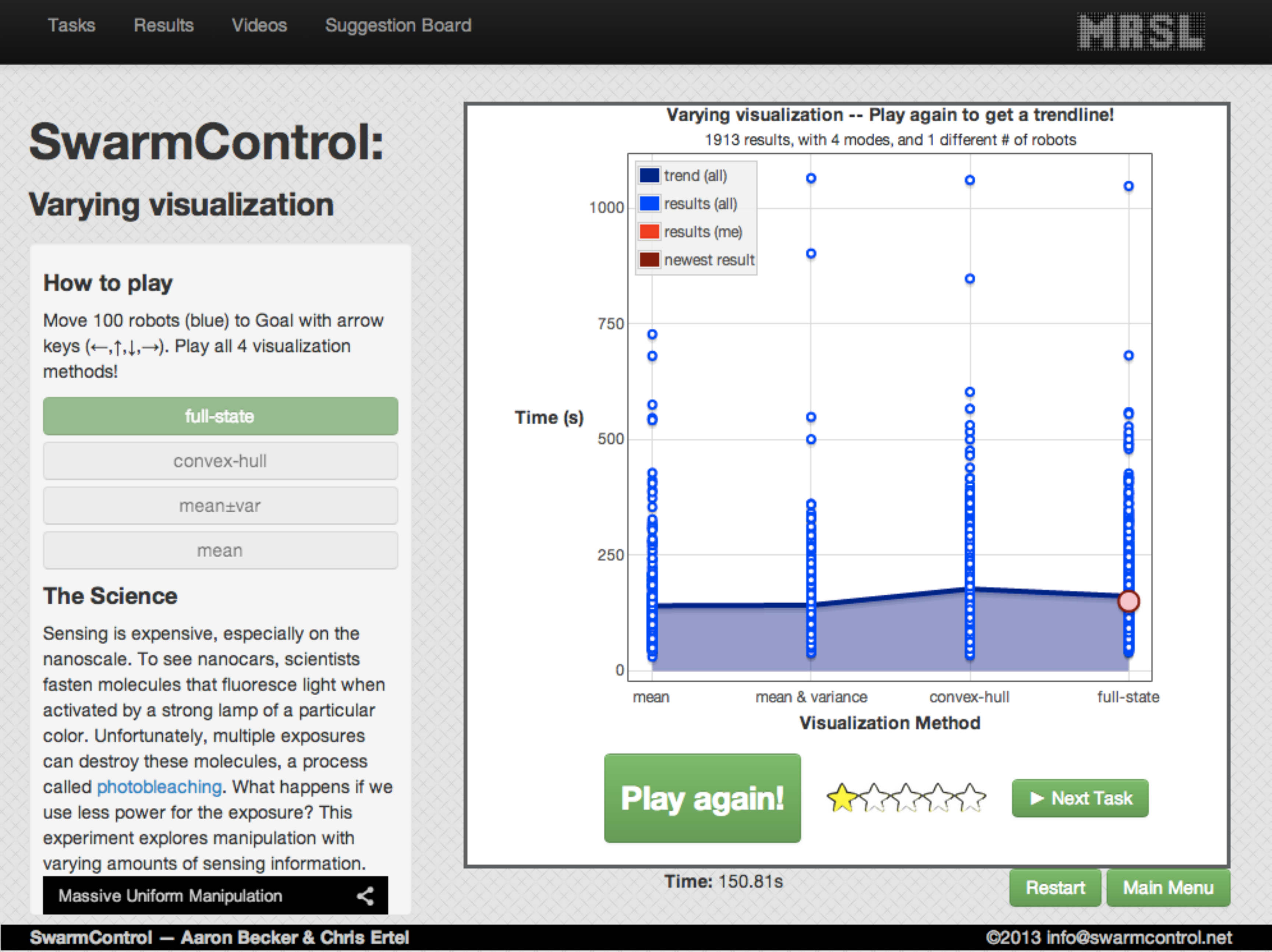}\end{overpic}
\vspace{-2em}
\caption{\label{fig:ScreenShotSuccess}Screenshot showing the \emph{Varying Visualization} experiment after a successful completion. Merit badges, feedback with comparison to other players, and a large \emph{Play again!} button contribute to encourage multiple experiments.
\vspace{-2em}
}
\end{figure}

To signal completion of a task and to show participants how they had performed we added a results screen to the end of each experiment. When a user completes an experiment trial, they are congratulated, shown their completion time, and shown how their results compare with everyone else. Each experiment has a number of blank merit badge outlines showing how many trials the participant should complete. Finishing an experiment fills in one of these outlines. These additions are shown in Fig.~\ref{fig:ScreenShotSuccess}.

To explain why the experiment matters, we added  \emph{The Science} sections on each page. These sections gave practical context to the experiment being performed, showing its utility in real-world situations.

\subsubsection{Respect participant time}
For experiments conducted online, it is important to waste as little of a participant's time as possible: you are in direct competition with a thousand other distractions. Our beta testers were frustrated by slow robots; in contrast, when we have conducted in-person tests with hardware robots in the lab, people have been more patient. To fix this problem we sped up our simulated robots by a factor of five.

\subsubsection{Test multiple displays}

Finally, many participants used laptop screens---displays much smaller than the desktops on which we designed the experiments. We changed our framework so that laptop participants could play without scrolling the screen.

\subsection{Lessons learned: Live website}

\begin{figure}
\begin{overpic}[width = \columnwidth]{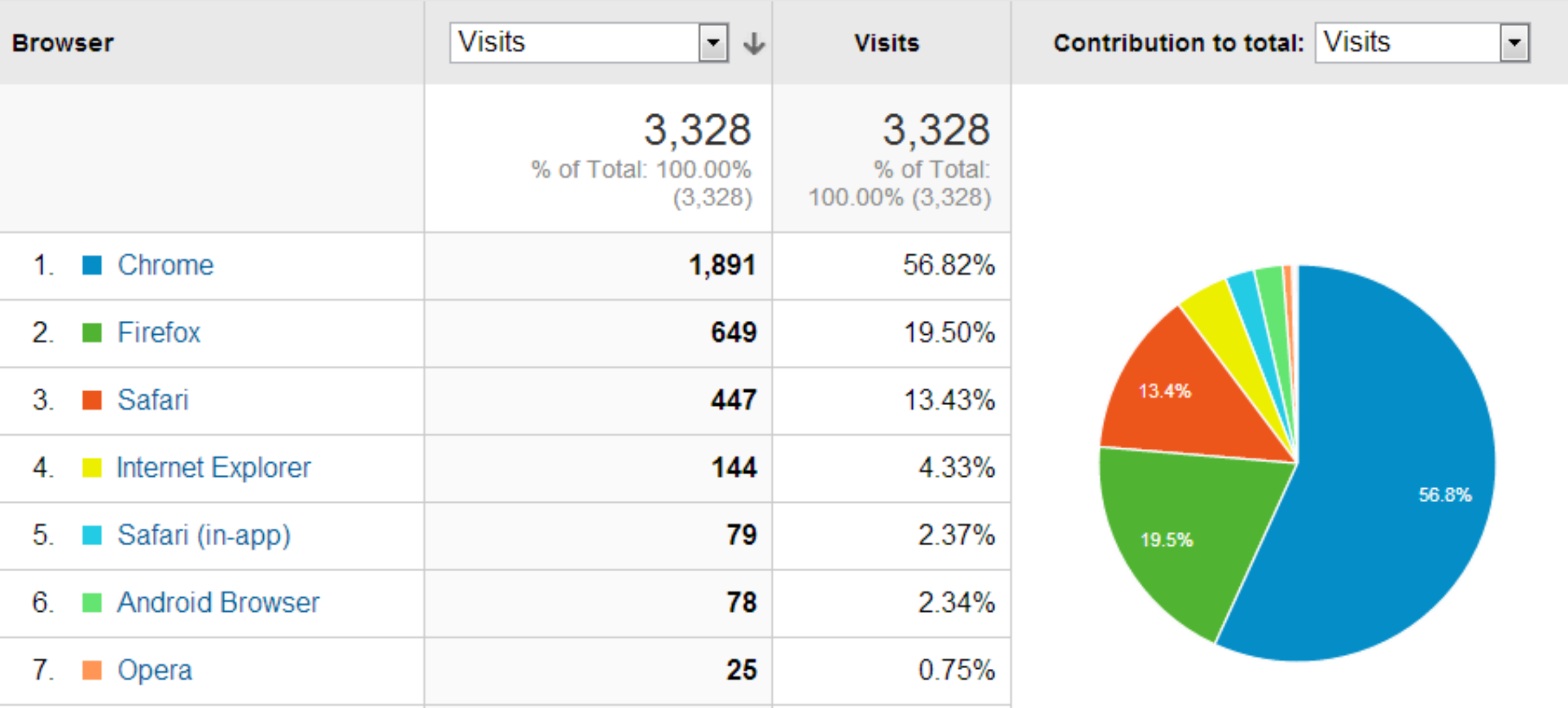}\end{overpic}
\vspace{-2em}
\caption{\label{fig:Browsers}Overview of the browsers used by game players, provided by Google Analytics. 
\vspace{-2em}
}
\end{figure}

After beta-testing, we launched the website and monitored our traffic using saved results and tracking by Google Analytics. This information was helpful in finding several usage trends.

\subsubsection{Mobile traffic}
 Looking at this data, we found that visitors using a smartphone or tablet leave within 17 seconds on average. These participants are 10\% of our first-time visitors, and leave because our experiments require the use of a mouse and keyboard. To capture these participants, we plan to make a mobile-friendly version of our experiments.

\subsubsection{Browser considerations}
Browser information for our experiments is shown in Fig.~\ref{fig:Browsers}.
For an online experiment, the experimenter cannot control the participant's browser. This imposes challenges for controlled experiments.

A large screen and a small screen---unless designed for---may show content differently, leading to inconsistent results. We optimized for a small laptop screen.

Additionally, certain browser/computer combinations will exhibit abysmal performance as the number of robots increases. Performance may suffer because of other processes the participant is running. Two ways to compensate for variance in performance are to conduct large numbers of experiments or to record performance data concurrently with experiment data. We attempted both: after one week of testing we began recording the user-agent string of participants' browsers. A better setup would benchmark a participant's browser before running an experiment.

%current website optimized for research, not for education.

\subsubsection{Bounce rates}
Participants are easily put off by having to click through or sign-up for things~\cite{krug2009don}.  Our bounce rate of 42\% and average visit time of just over 4 minutes indicate that almost half potential testers never attempt an experiment. We saw an exponential decrease in the number of trials participants completed, as shown in Fig.~\ref{fig:Learning}.

We originally planned on a participant sign-up process, but omitted it to let participants begin experiments faster. A participant can start running an experiment now in two clicks.

%https://www.google.com/analytics/web/?hl=en&pli=1#report/visitors-overview/a42817804w72642836p74969041/

\subsection{Other issues}

There are ways to improve the quality of our experimentation.

\subsubsection{Poor experiment design}

We tried to design each experiment to measure the effect one parameter; we did not always succeed. In particular, the \emph{Position Control} experiment uses varying numbers of robots---however, there is no difference in input required to solve certain configurations. Identical input sequences solve $n$=1 and $n$=2, a problem also seen with $n$=3 and $n$=4.

Additionally, there is a qualitative difference in difficulty when arranging the robots into a solid versus hollow shape; this is currently not controlled for. A possible solution would require goal states for all values of $n$ to be either solid or hollow.

\subsubsection{Missing participant behavior}

We only recorded results when a participant successfully completed a task---we did not track the case where a participant began but did not complete an experiment. It is important to account for every path a participant may take through an experiment. We did not, and found a discrepancy between the visits reported by Google Analytics and the number of experiment results recorded.

%%%%%%%%%%%%%%%
%%%%%%%%%%%%%%%
%%%%%%%%%%%%%%%%%%%%%%%%%%%%%%%%%%%%%%%%%%%%%%%%%%%%%%%%%%%
\section{Conclusion and Future Work}\label{sec:conclusion}
%%%%%%%%%%%%%%%%%%%%%%%%%%%%%%%%%%%%%%%%%%%%%%%%%%%%%%%%%%%
    
\begin{figure}
\begin{overpic}[width = 0.48\columnwidth]{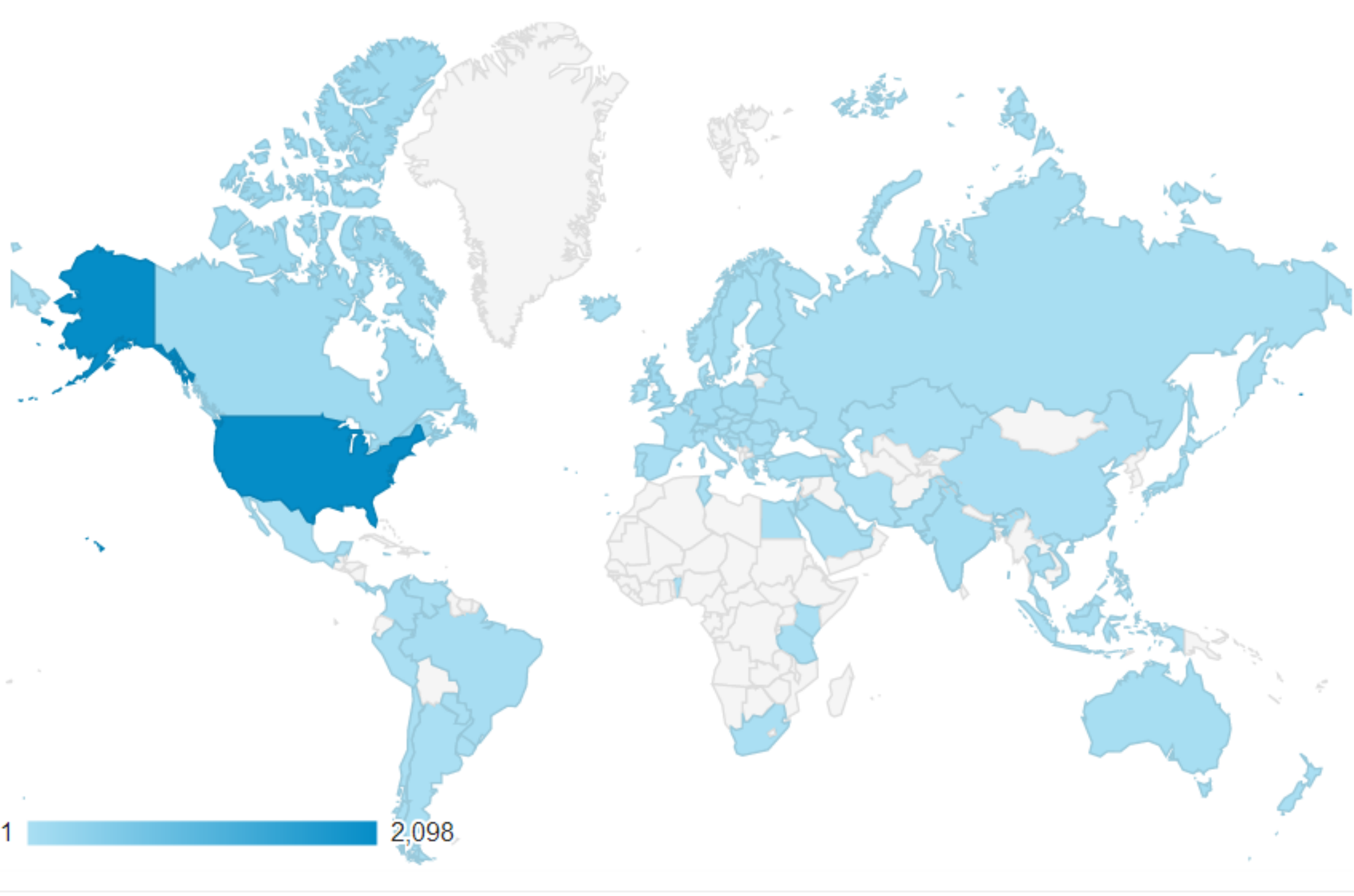}\end{overpic}
\begin{overpic}[width = 0.48\columnwidth]{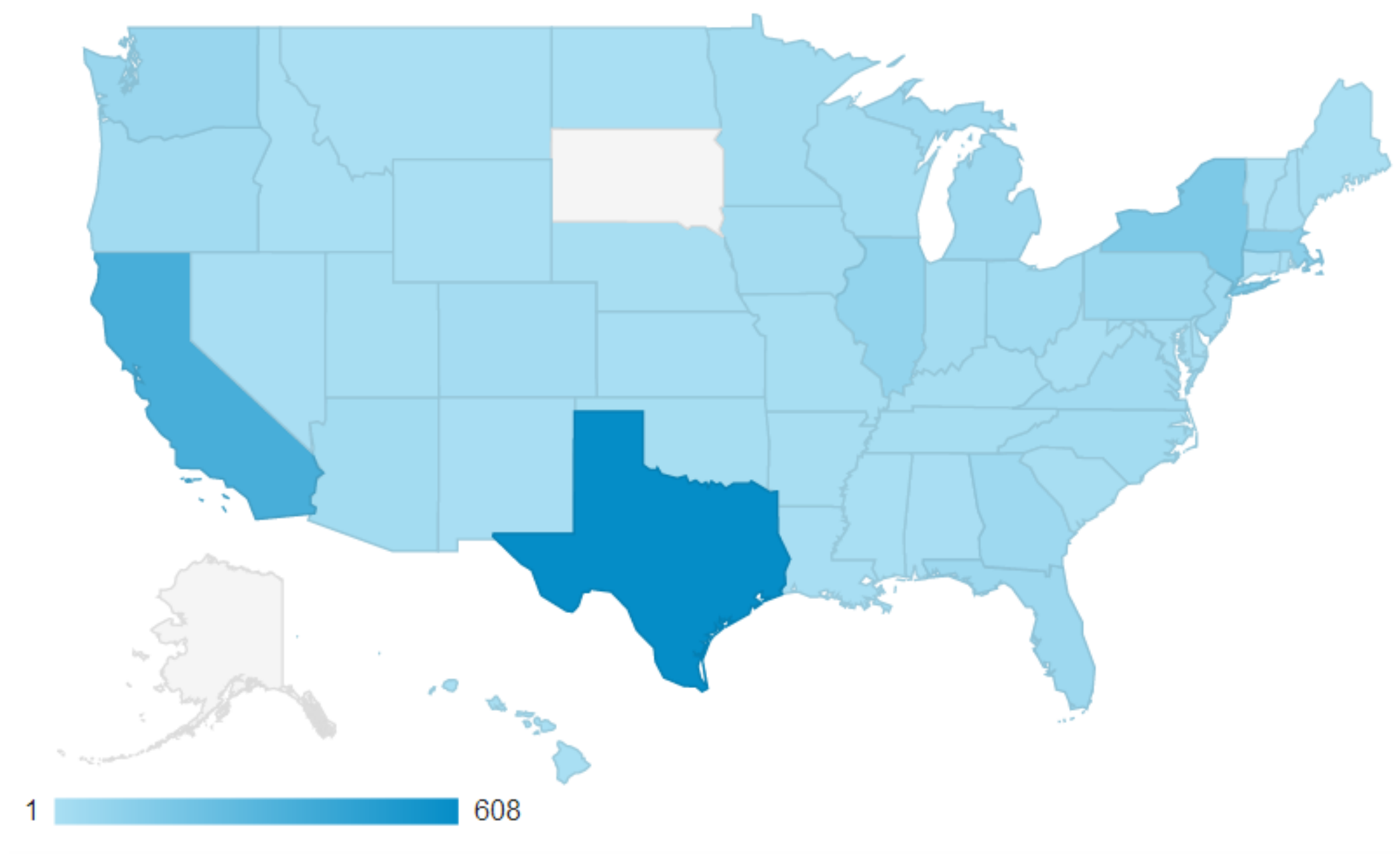}\end{overpic}
\vspace{-1em}
\caption{\label{fig:PlayerLocation}Demographic information on game player's location, provided by Google Analytics. Game players from 84 countries and 49 US states visited our site.
\vspace{-2em}
}
\end{figure}

We introduced \href{http://www.swarmcontrol.net/}{SwarmControl.net}, a new online environment for large-scale user experiments controlling 100+ populations of robots.  Over the period of one month this site conducted thousands of experiments with a worldwide user base, as shown in Fig.~\ref{fig:PlayerLocation}.   \href{https://github.com/crertel/swarmmanipulate.git}{All code is open source and downloadable from a public git repository}~\cite{Chris-Ertel2013}. \href{http://www.swarmcontrol.net/show_results}{All experiment results can be freely downloaded from our website}.  We implemented five unique experiments, and gathered data that corroborated past lab experiments, but with a testing pool two orders of magnitude larger than was possible before.

\paragraph{Site modifications}
  The current site is optimized for desktop and laptop users, and we currently do not support mobile users. Our IRB allows us to conduct demographic questionnaires, and we will implement these questionnaires in a future release--currently our only source of demographic data is Google Analytics.
  
  We are pursuing partnerships to increase the educational content on our website. Our goal is to highlight a variety of the leading micro- and nano-robotics labs and the challenges they are working on.

\paragraph{Additional experiments}
There are many avenues for future work.  Manipulation by large populations of robots is an immature area and there are many open questions. Future work will invite other collaborators to submit their own experiments.
Topics of interest include  control with nonuniform flow such as fluid in an artery, gradient control fields like that of an MRI, competitive playing, multi-modal control, and targeted drug delivery in a vascular network.

\paragraph{Automatic controllers}
We have compiled a large body of test results.  Our goal is to design automatic controllers using this data. One avenue is to identify the most proficient players and perform inverse optimal control algorithms to learn the cost functions used by the best players.  

\section{Acknowledgements}
We acknowledge many \href{http://mrsl.rice.edu/}{mrsl.rice.edu} lab members who helped beta test \href{http://www.swarmcontrol.net}{SwarmControl.net}; helpful discussion with \href{http://www.jmtour.com/}{James Tour}, \href{http://slink.rice.edu/}{Stephan Link}, and Victor Garc\`ia L\`opez on fabrication and visualization challenges at the nanoscale with nanocars; and the \href{http://cttl.rice.edu/}{Rice Center for Technology in Teaching and Learning}.
This work was partially supported by the National Science Foundation under 
\href{http://www.nsf.gov/awardsearch/showAward?AWD_ID=1035716}{CPS-1035716}.  
   
\bibliographystyle{IEEEtran}
\bibliography{IEEEabrv,aaronrefs}%,../../../ensemble/bib/aaronrefs}%,../aaronrefs}
\end{document}